\documentclass[sigconf]{acmart}
\AtBeginDocument{%
  \providecommand\BibTeX{{%
    \normalfont B\kern-0.5em{\scshape i\kern-0.25em b}\kern-0.8em\TeX}}}

\copyrightyear{2023}
\acmYear{2023}
\setcopyright{acmlicensed}
\acmConference[KDD '23]{Proceedings of the 29th ACM SIGKDD Conference on Knowledge Discovery and Data Mining}{August 6--10, 2023}{Long Beach, CA, USA}
\acmBooktitle{Proceedings of the 29th ACM SIGKDD Conference on Knowledge Discovery and Data Mining (KDD '23), August 6--10, 2023, Long Beach, CA, USA}
\acmPrice{15.00}
\acmISBN{979-8-4007-0103-0/23/08}
\acmDOI{10.1145/3580305.3599548} 
\usepackage{enumitem}
 \usepackage{bm}
 \usepackage{balance} 
 \usepackage{amsmath}
\usepackage{svg}
\usepackage{arydshln}
\usepackage{multirow}
\usepackage{geometry}%页面设置
\usepackage{graphics}%图片设置
\usepackage{caption}%注释设置
\usepackage{threeparttable}
\usepackage{graphicx}
\usepackage{wrapfig}
\usepackage{xcolor}  
\usepackage{tabularx}
\usepackage{booktabs,multirow,diagbox}
\usepackage{float}
\usepackage{subfig}
\usepackage{algorithm,algorithmic}
\usepackage{stmaryrd}
\definecolor{ggray}{HTML}{E7E6E6}
\usepackage{colortbl}

\newtheorem{definition}{Definition}

\newtheorem{theorem}{Theorem}[section]

\newtheorem{lemma}{Lemma}

\newcommand{\hide}[1]{} %hide

\newcommand{\vpara}[1]{\vspace{0.05in}\noindent\textbf{#1 }}
\begin{document}

\title[When to Pre-Train Graph Neural Networks? From Data Generation Perspective!]{When to Pre-Train Graph Neural Networks? \\ From Data Generation Perspective!}

\author{Yuxuan Cao}
\affiliation{
  \institution{Zhejiang University, Fudan University}
   \country{}
}
% \affiliation{%
%   \institution{Fudan University}
%   % \country{Shanghai, China}
% }
\authornote{Both authors contributed equally to this research. }
\authornote{This work was done when the author was a visiting student at Fudan University.}
\email{caoyx@zju.edu.cn}

\author{Jiarong Xu}
\affiliation{%
  \institution{Fudan University}
   \country{}
}
\authornotemark[1]
\authornote{Corresponding author.}
 \email{ jiarongxu@fudan.edu.cn}
 
\author{Carl Yang}
\affiliation{%
  \institution{Emory University}
  \country{}
}
\email{j.carlyang@emory.edu}
\author{Jiaan Wang}
\affiliation{%
  \institution{Soochow University}
  \country{}
}
 \email{jawang.nlp@gmail.com}
\author{Yunchao Zhang}
\affiliation{%
  \institution{Zhejiang University}
\country{}
}
 \email{m.yunchaozhang@gmail.com}
\author{Chunping Wang}
\affiliation{%
  \institution{Finvolution Group}
  \country{}
}
 \email{ wangchunping02@xinye.com}
\author{Lei Chen}
\affiliation{%
  \institution{Finvolution Group}
  \country{}
}
 \email{ chenlei04@xinye.com}
 \author{Yang Yang}
\affiliation{%
  \institution{Zhejiang University}
  \country{}
  }
 \email{ yangya@zju.edu.cn}

\renewcommand{\shortauthors}{Cao and Xu, et al.}
\begin{abstract}
\noindent In recent years, graph pre-training has gained significant attention, focusing on acquiring transferable knowledge from unlabeled graph data to improve downstream 
performance. 
Despite these recent endeavors, the problem of negative transfer remains a major concern when utilizing graph pre-trained models to downstream tasks. Previous studies made great efforts on the issue of \emph{what to pre-train} and \emph{how to pre-train} by designing a variety of graph pre-training and fine-tuning strategies. However, there are cases where even the most advanced ``pre-train and fine-tune'' paradigms fail to yield distinct benefits.
This paper introduces a generic framework W2PGNN to answer the crucial question of \emph{when to pre-train} (\emph{i.e.}, in what situations could we take advantage of graph pre-training) before performing effortful pre-training or fine-tuning. We start from a new perspective to explore the complex generative mechanisms from the pre-training data to downstream data. In particular, W2PGNN first fits the pre-training data into graphon bases, each element of graphon basis (\emph{i.e.}, a graphon) identifies a fundamental transferable pattern shared by a collection of pre-training graphs. All convex combinations of graphon bases give rise to a generator space, from which graphs generated form the solution space for those downstream data that can benefit from pre-training. In this manner, the feasibility of pre-training can be quantified as the generation probability of the downstream data from any generator in the generator space. W2PGNN offers three broad applications: providing the application scope of graph pre-trained models, quantifying the feasibility of pre-training, and assistance in selecting pre-training data to enhance downstream performance. We provide a theoretically sound solution for the first application and extensive empirical justifications for the latter two applications.

\end{abstract}


\maketitle

\section{Introduction}
Graph neural networks (GNNs) have undergone rapid development and become increasingly popular for learning graph data \cite{welling2016semi, velivckovic2017graph, xu2018powerful}.
GNNs are usually trained in an end-to-end manner while getting enough labeled data is arduously expensive and sometimes even impractical to access. This motivates some recent advances in pre-training GNNs ~\cite{hu2019strategies,Hu2020GPTGNNGP,Qiu2020GCCGC,Lu2021LearningTP, liu2022user}. 
The key insight of pre-training GNNs is to learn transferable knowledge from a collection of unlabeled graph data, hoping that the learned knowledge can be easily adapted to downstream tasks.
In view of the great success of pre-training in other fields like computer vision and natural language processing~\cite{devlin2018bert,he2020momentum},  graph pre-training is {highly expected} to be an effective means to improve downstream performance.

\begin{figure}[t]
    \centering
    {\includegraphics[width=1\columnwidth]{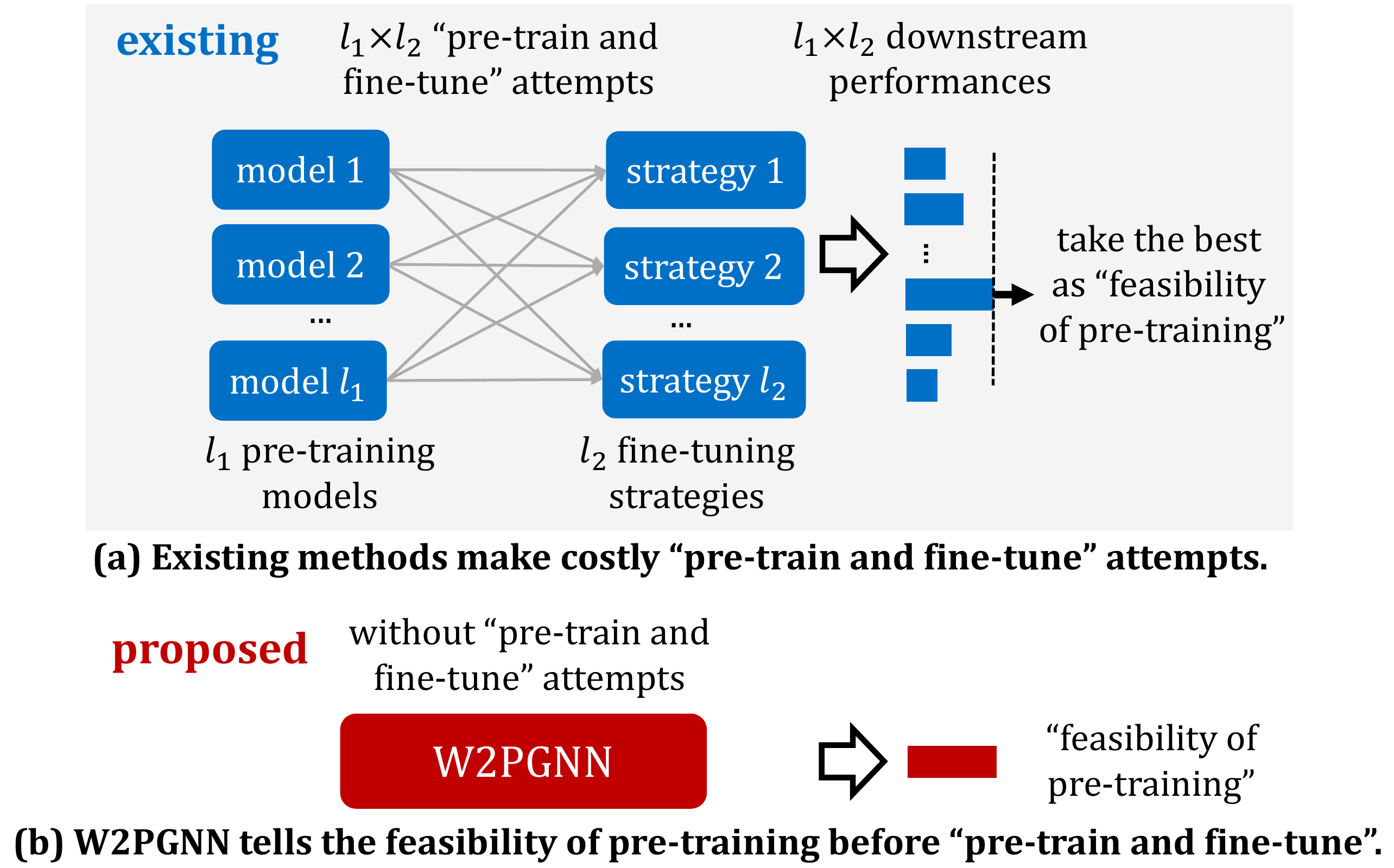}}
    \caption{Comparison of {existing methods} and {proposed W2PGNN} to answer \emph{when to pre-train} GNNs.}    
    \label{fig:example}
    \vspace{-0.1in}
\end{figure}

However, the intuition that graph pre-trained model would ideally benefit the downstream is far from the truth in the area of graph pre-training.
Instead, graph pre-trained models can lead to \emph{negative transfer} on many downstream tasks, especially when the graphs used for pre-training are not necessarily from the same domain as the {downstream} data~\cite{hu2019strategies, Qiu2020GCCGC}.
For example, the closed triangles ($\vcenter{\hbox{\includegraphics[width=2.4ex,height=2.4ex]{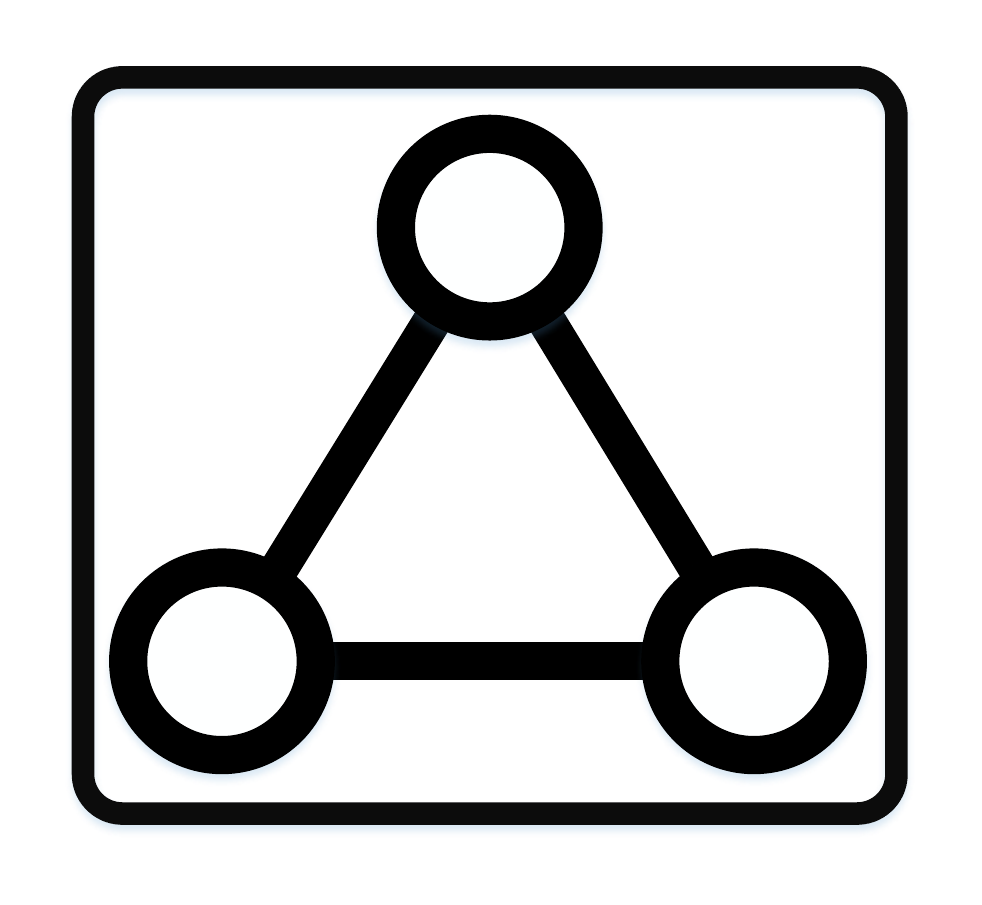}}}$) and open triangles  ($\vcenter{\hbox{\includegraphics[width=2.4ex,height=2.4ex]{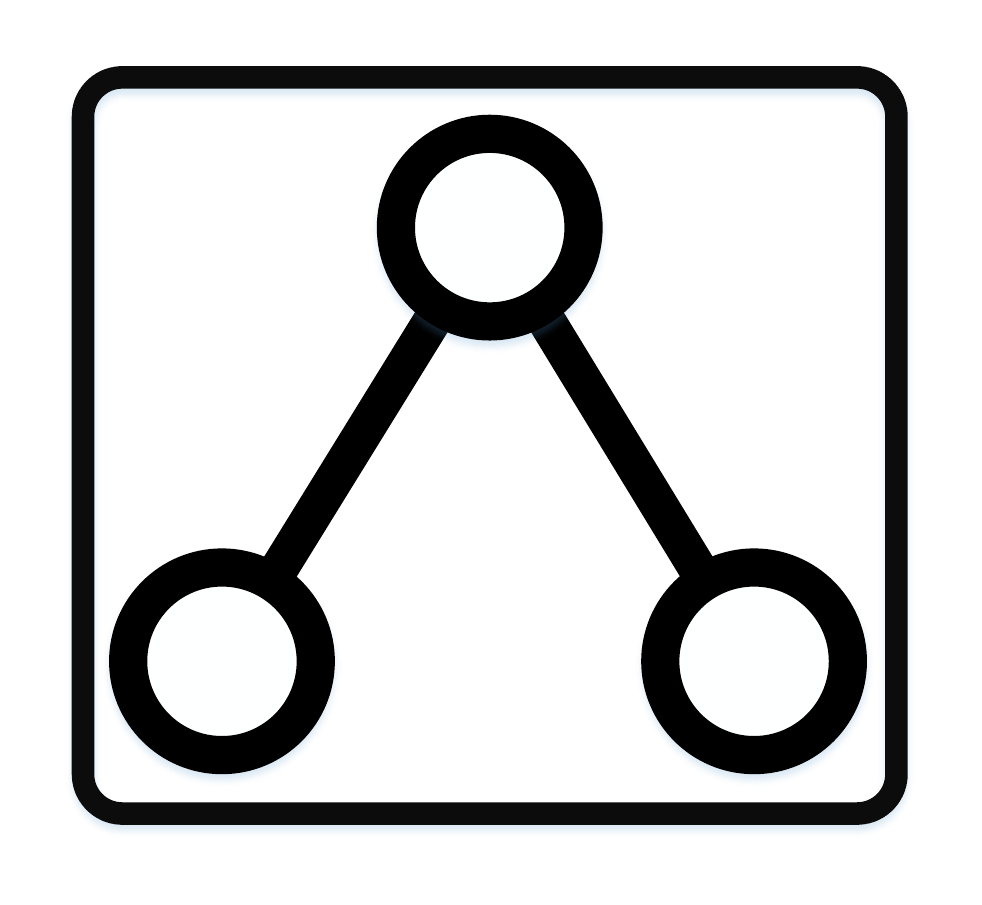}}}$) might yield different interpretations in molecular networks (unstable vs. stable in terms of chemical property) from those in social networks (stable vs. unstable in terms of social relationship); such distinct or reversed semantics does not contribute to transferability, and even exacerbates the problem of negative transfer.

To avoid the negative transfer, recent efforts focus on  \emph{what to pre-train} and \emph{how to pre-train},  \emph{i.e.}, design/adopt graph pre-training models with a variety of self-supervised tasks to capture different patterns~\cite{Qiu2020GCCGC,you2020graph,Lu2021LearningTP} and fine-tuning strategies to enhance downstream performance~\cite{Hu2019PreTrainingGN,Han2021AdaptiveTL,Zhang2022FineTuningGN,Xia2022TowardsEA}.
However, there do exist some cases that no matter how advanced the pre-training/fine-tuning method is, the transferability from pre-training data to downstream data still cannot be guaranteed. This is because the underlying assumption of deep learning models is that the test data should share a similar distribution as the training data.
Therefore, it is a necessity to understand \emph{when to pre-train}, \emph{i.e.}, under what situations the ``graph pre-train and fine-tune'' paradigm should be adopted.

Towards the answer of when to pre-train GNNs, one straight-forward way illustrated in Figure~\ref{fig:example}(a) is to train and evaluate on all candidates of pre-training models and fine-tuning strategies, and then the resulting best downstream performance would tell us whether pre-training
% ``pre-train and fine-tune'' 
is a sensible choice. If there exist $l_1$ pre-training models and $l_2$ fine-tuning strategies,  such a process would be very costly as you should make $l_1 \times l_2$ ``pre-train and fine-tune'' attempts.
Another approach is to utilize graph metrics to measure the similarity between pre-training and downstream data, \emph{e.g.}, density, clustering coefficient and etc. However, it is a daunting task to enumerate all hand-engineered graph features or find the dominant features that influenced similarity.
Moreover, the graph metrics only measure the pair-wise similarity between two graphs, which cannot be directly and accurately applied to the practical scenario where pre-training data contains multiple graphs.

In this paper, we propose a W2PGNN framework to answer
\emph{\underline{w}hen \underline{to} \underline{p}re-train \underline{GNN}s from a graph data generation perspective}.
% aim to address the problem of when to pre-train GNNs 
The high-level idea is that instead of performing effortful graph pre-training/fine-tuning or making comparisons between the pre-training and downstream data, we study the complex generative mechanism from the pre-training data to the downstream data (Figure~\ref{fig:example}(b)).
We say that downstream data can benefit from pre-training data (\emph{i.e.}, has high feasibility of performing pre-training), 
if it can be generated with high probability by a graph generator that summarizes the topological characteristic of pre-training data.

The major challenge is how to obtain an appropriate graph generator, hoping that it not only inherits the transferable topological patterns of the pre-training data, but also is endowed with the ability to generate feasible downstream graphs.
To tackle the challenge, we propose to design a graph generator based on graphons.
We first fit the pre-training graphs into different graphons to construct a \emph{graphon basis}, where each graphon (\emph{i.e.}, element of the graphon basis) identifies a collection of graphs that share common transferable patterns. We then define a \emph{graph generator} as {a convex combination of elements in a graphon basis}, which serves as a comprehensive and representative summary of pre-training data.  All of these possible generators constitute the \emph{generator space}, from which graphs generated form the solution space for the downstream data that can benefit from pre-training.

Accordingly, the feasibility of performing pre-training can be measured as the highest probability of downstream data being generated from any graph generator in the generator space, which can be formulated as an optimization problem.
However, this problem is still difficult to solve due to the large search space of graphon basis. We propose to reduce the search space to three candidates of graphon basis, \emph{i.e.,} topological graphon basis, domain graphon basis, and integrated graphon basis, to mimic different {generation mechanisms} from pre-training to downstream data. Built upon the reduced search space, the feasibility can be approximated efficiently.

Our major contributions are concluded as follows:
\begin{itemize}[leftmargin=*,topsep=0pt]
\item \textbf{Problem and method.} To the best of our knowledge, we are the first work to study the problem of when to pre-train GNNs. We propose a W2PGNN framework
to answer the question from a data generation perspective, which tells us the feasibility of performing graph pre-training before conducting effortful pre-training and fine-tuning.

\item \textbf{Broad applications.}
W2PGNN provides several practical applications: (1) provide the application scope of a graph pre-trained model, (2) measure the feasibility of performing pre-training for a downstream data
and (3) choose the pre-training data so as to maximize downstream performance with limited resources.

\item \textbf{Theory and Experiment.} 
We theoretically and empirically justify the effectiveness of W2PGNN.
Extensive experiments {on real-world graph datasets from multiple domains} show that the proposed method can provide an accurate estimation of pre-training feasibility and the selected pre-training data can benefit the downstream performance.

\end{itemize}

\section{Problem Formulation}\label{sec:problem}

In this section, we first formally define the problem of when to pre-train GNNs. Then, we provide a brief theoretical analysis of the transferable patterns in the problem we study, and finally discuss some non-transferable patterns.

\begin{definition}[When to pre-train GNNs]
    Given the pretraining graph data $\mathcal G_\text{train}$ and the downstream graph data $\mathcal G_\text{down}$, our main goal is to answer to what extent the ``pre-train and fine-tune'' paradigm can benefit the downstream data.
\end{definition}
Note that in addition to this main problem, our proposed framework can also serve other scenarios, such as providing the application scope of graph pre-trained models, and helping select pre-training data to benefit the downstream (please refer to the \emph{application cases} in Section~\ref{subsec:overview} for details).

\vpara{Transferable graph patterns.}
The success of ``pre-train and fine-tune'' paradigm is typically attributed to the commonplace between pre-training and downstream data. However, in real-world scenarios, there possibly exists a significant divergence between the pre-training data and the downstream data. To answer the problem of when to pre-train GNNs, the primary task is to define the transferable patterns across graphs.

We here theoretically explore which patterns are transferable between pre-training and downstream data under the performance guarantee of graph pre-training model (with GNN as the backbone).
\begin{theorem}[Transferability of graph pre-training model]\label{thero:emb}
Let $G_\text{train}$ and $G_\text{down}$ be two (sub)graphs sampled from $\mathcal G_\text{train}$ and $\mathcal{G}_\text{down}$, and assume the  attribute of each node as a scalar $1$ without loss of generality. Given a graph pre-training model $e$ (instantiated as a GNN)  with $K$ layers and $1-$hop graph filter $ \Phi(L)$ (which is a function of the normalized graph Laplacian matrix $L$), we have 

\begin{equation}
\left\|e(G_\text{train})-e(G_\text{down})\right\|_2
\leq \kappa
\Delta_\text{topo}\left(G_\text{train}, G_\text{down}\right)
\end{equation}
where {$\Delta_\text{topo}\left(G_\text{train}, G_\text{down}\right) =  \frac{1}{mn}  \sum_{i=1}^m \sum_{j^{\prime}=1}^n \|
L_{g_i}-L_{g_j^{\prime}} \|_2$} measures the topological divergence between $G_\text{train}$ and $G_\text{down}$, where
$g_i$ is the $K$-hop ego-network of node $i$ from $G_\text{train}$ and $L_{g_i}$ is its corresponding normalized graph Laplacian matrix, $m$ and $n$ are the number of nodes of $G_\text{train}$ and $G_\text{down}$. $e(G_\text{train})$ and $e(G_\text{down})$ are the output representations of $G_\text{train}$ and $G_\text{down}$ from graph pre-training model,
$\kappa$ is a constant relevant to  $K$, graph filter $\Phi$, learnable parameters of GNN and the activation function used in GNN.
\end{theorem}

% Detailed proofs and descriptions can be found in Appendix~\ref{proof:emb}.
Theorem~\ref{thero:emb} suggests that two (sub)graphs sampled from pre-training and downstream data with similar topology are transferable via graph pre-training model (\emph{i.e.}, sharing similar representations produced by the model).
{Hence we consider the transferable graph pattern as the topology of a (sub)graph, either node-level or graph-level. Specifically, the node-level transferable pattern could be the topology of the ego-network of a node (or the structural role of a node), irrespective of the node's exact location in the graph. The graph-level transferable pattern is the topology of the entire graph itself (\emph{e.g.}, molecular network). Such transferable patterns constitute the input space introduced in Section~\ref{subsec:overview}. }

\vpara{Discussion of non-transferable graph patterns.}
As a remark, we show that two important pieces of information (\emph{i.e.}, attributes and proximity) commonly used in graph learning are not necessarily transferable across pre-training and downstream data in most real-world scenarios, thus we do not discuss them in this paper.

First, although the attributes carry important semantic meaning in one graph, 
it can be shown that the attribute space of different graphs typically has little or no overlap at all. For example, if the pre-training and downstream data come from different domains, their nodes would indicate different types of entities and the corresponding attributes may be completely irrelevant. Even for graphs from the similar/same domain, the dimensions/meaning of their node attributes can be totally different and result in misalignment. 

The proximity, on the other hand, 
assumes that closely connected nodes are similar,
which also cannot be transferred across graphs. 
This assumption depends on the overlaps in neighborhoods and thus only works on graphs with the same or overlapped node set.

\section{Preliminary and Related Works}
\vspace{-0.05in}
\vpara{Graphons.}
A Graphon (short for graph function)~\cite{Airoldi2013StochasticBA} is a bounded symmetric function $B: [0,1]^{2} \rightarrow [0,1]$ (different subscripts of $B$ denote different graphons), which can be interpreted as the weighted matrix of an arbitrary undirected graph with uncountable number of nodes\cite{lovasz2012large}. Literaturelly, graphon has been studied from two perspectives: as limit of graph sequence, and as graph generators\cite{lovasz2006limits,Airoldi2013StochasticBA, Han2022GMixupGD}. \emph{We utilize both perspectives in our framework.}

On one hand, a graphon can be considered as the limit objects of graph sequence,
% where the density of certain ``graph motifs'' can be preserved;
and every convergent graph sequence would converge to a graphon\cite{lovasz2012large}.
Thus, a graphon is a comprehensive summary of a collection of arbitrary size graphs. These graphs can be considered topologically similar in the sense that they belong to the same graphon. \emph{In this paper, we utilize a set of graphons as a comprehensive and representative summary of pre-training data.}

Taking graphon 
as a graph generator, we can associate nodes $i$ and $j$ with points $v_i$ and $v_j$ in $[0,1]$, and then $B(v_i , v_j)$ serves as the probability to generate the edge between these two nodes. Therefore, a graphon $B$ can generate unweighted graphs of arbitrary sizes, which can be taken as those induced graphs  potentially inheriting the topological patterns implied in graphon. \emph{Thus, the generation capability of graphon can help us  generate feasible downstream graphs that can benefit from pre-training.}

\vpara{Graph pre-training and fine-tuning.}
Graph pre-training models first learn universal knowledge from large-scale graph datasets with self-supervised or unsupervised objectives (\emph{i.e.}, pre-training stage), and then transfer the knowledge to deal with specific downstream tasks (\emph{i.e.}, fine-tuning stage).
Among them, some researchers design pre-training tasks based on the neighborhood similarity assumption~\cite{Perozzi2014DeepWalkOL,Kipf2016VariationalGA,Sun2020MultiStageSL,Grover2016node2vecSF,Donnat2018LearningSN,Zhang2019ProNEFA,Tang2015LINELI,Zhao2021DataAF}. The learned graph-specific patterns/knowledge could benefit downstream tasks on the same graphs, but cannot be generalized to unseen graphs.
To enhance the transferability of the pre-trained models, some works try to utilize graph data from the same (or similar) domains as the pre-training data~\cite{Hafidi2020GraphCLCS,hassani2020contrastive,Sun2020InfoGraphUA,Narayanan2017graph2vecLD,Zhu2020DeepGC,hu2019strategies,you2020graph,You2020WhenDS,Hu2020GPTGNNGP,Li2021PairwiseHD,Lu2021LearningTP,Sun2021MoCLDM,Zhang2021MotifbasedGS}, or explore cross-domain pre-training strategies~\cite{Qiu2020GCCGC,you2020graph,Lu2021LearningTP} as well as fine-tuning strategies~\cite{Hu2019PreTrainingGN,Han2021AdaptiveTL,Zhang2022FineTuningGN,Xia2022TowardsEA}.
Nevertheless, all these efforts focus on addressing the problem of  \emph{what to pre-train and how to pre-train} by developing pre-training or fine-tuning methods.
To the first time, we aim to study \emph{when to pre-train} GNNs, \emph{i.e.}, in what situations the graph pre-training should be adopted.

% \vpara{Transferability measure.}
% There are several attempts to measure the transferability of GNNs. 
% The most  straightforword way is to train and evaluate on 
% all candidates of pre-training models and fine-tuning strategies,
% and then the resulting best downstream performance as the transferability measure.
% However, as depicted in Figure~\ref{fig:example}(a), such approach would be very costly to perform effortful pre-training and fine-tuning. 
% Another way is based on graph properties, which leverage the graph  properties (\emph{e.g.,} degree~\cite{borner2007network}, density~\cite{wasserman1994social}, assortativity~\cite{newman2003mixing} and etc.) to measure the similarities between pre-training and downstream graphs, potentially can be utilized to approximate the transferability. 
% Some other works also focus on analyzing the transferability of GNNs theoritically~\cite{levie2021transferability,Ruiz2021TransferabilityPO}. Nevertheless, they are limited to measure the transferability of GNNs on a single graph or when training and testing data are from the same dataset~\cite{levie2021transferability,Ruiz2021TransferabilityPO}  , which are inapplicable to our setting.
% A recent work, EGI~\cite{Zhu2021TransferLO}  addresses the transferability measure problem of GNNs across graphs. However, EGI is a model-specific measure and depend on its own framework. 
% For the first time, we study the transferability of graph pre-training from the data perspective, without performing any pre-training and fine-tuning.

% \vspace{-0.05in}
\section{Methodology}
In this section, we first present our proposed framework W2PGNN to answer when to pre-train GNNs in Section~\ref{subsec:overview}.
Based on the framework, we further introduce the measure of
the feasibility of performing pre-training in Section~\ref{subsec:fea}.
Then in Section~\ref{subsec:3.3}, we discuss our approximation to the feasibility of pre-training.
Finally, the complexity analysis of W2PGNN is provided in Section~\ref{subsec:cost}.

\begin{figure}[t]
\centerline{\includegraphics[width=0.9\linewidth]{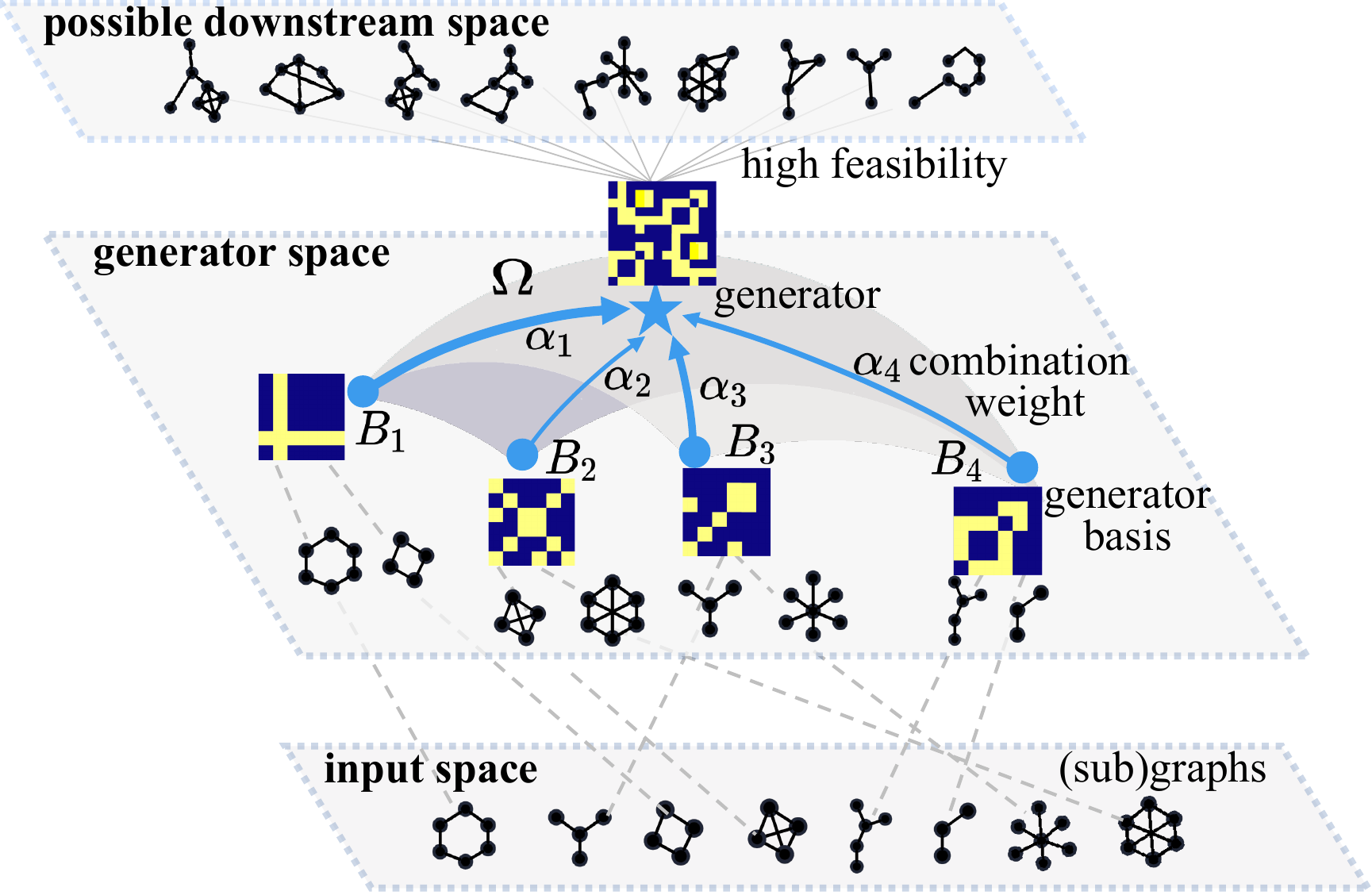}} 
\caption{Illustration of our proposed framework W2PGNN to answer when to pre-train GNNs.}
\vspace{-0.1in}
\label{fig:model}
\end{figure}
\vspace{-0.05in}
\subsection{Framework Overview}
\label{subsec:overview}

 W2PGNN framework provides a guide to answer
\emph{\underline{w}hen \underline{to} \underline{p}re-train \underline{GNN}s from a data generation perspective}.
The key insight is that if downstream data can be generated with high probability by a graph generator that summarizes the pre-training data, the downstream data would
 present high feasibility of performing pre-training.

The overall framework of W2PGNN can be found in Figure~\ref{fig:model}.
Given the \emph{input space} consisting of pre-training graphs, we fit them into a graph generator in the \emph{generator space}, from which the graphs generated constitute the \emph{possible downstream space}.
More specifically,
an ideal graph generator should inherit different kinds of topological patterns, based on which new graphs can be induced. Therefore, we first construct a graphon basis $\mathcal{B}=\{B_1, B_2, \cdots, B_k\}$, where each element $B_i$ represents a graphon fitted from a set of (sub)graphs with similar patterns (\emph{i.e.}, the blue dots $\vcenter{\hbox{\includegraphics[width=2.1ex,height=2.1ex]{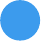}}}$). 
To access different combinations of generator basis,
each $B_i$ is assigned with a corresponding weight $\alpha_i$ (\emph{i.e.}, the width of blue arrow $\vcenter{\hbox{\includegraphics[width=2.3ex,height=2.1ex]{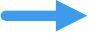}}}$) and their combination gives rise to a graph generator (\emph{i.e.}, the blue star $\vcenter{\hbox{\includegraphics[width=2.1ex,height=2.1ex]{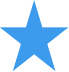}}}$). All weighted combinations compose the generator space $\Omega$ (\emph{i.e.}, the gray surface $\vcenter{\hbox{\includegraphics[width=3ex,height=2ex]{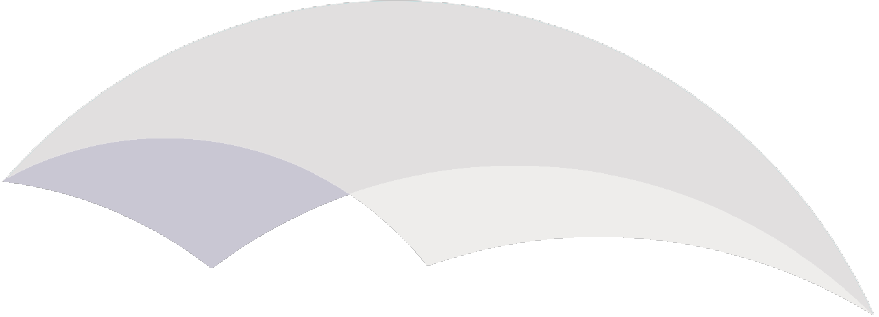}}}$), from which graphs generated form the possible solution space of downstream data (shorted as possible downstream space). 
{The generated graphs are those that could benefit from the pre-training data, we say that they exhibit \emph{high feasibility} of performing pre-training.}
 
In the following, we introduce the workflow of W2PGNN in the input space, the generator space and the possible downstream space in detail.
Then, the application cases of W2PGNN are given for different practical use.

\vpara{Input space.}
The input space of W2PGNN is composed of nodes' ego-networks or graphs. For node-level pre-training, we take the nodes' ego-networks to constitute the input space; For graph-level pre-training, we take the graphs  (\emph{e.g.}, small molecular graphs) as input space.

\vpara{Generator space.}
As illustrated in Figure~\ref{fig:model}, each point (\emph{i.e.}, graph generator) in the generator space $\Omega$ is a  convex combination of  generator basis $\mathcal{B}=\{B_1, B_2, \cdots, B_k\}$. Formally, we define the graph generator as
 % \vspace{-0.05in}
\begin{equation}
f(\{\alpha_i\},\{B_i\}) = \sum _{i=1}^k \alpha_i B_i, \ \ \text{where} \  \sum _{i=1}^k \alpha_i = 1, \alpha_i \geq 0.
% , 1 \leq i \leq k,
 \vspace{-0.05in}
\end{equation}
Different choices of $\{\alpha_i\},\{B_i\}$ comprise different graph generators.
All possible generators constitute the \emph{generator space}
$\Omega=\{ f(\{\alpha_i\},\{B_i\}) \mid \forall \ \{\alpha_i\},\{B_i\} \}$.

We shall also note that, the graph generator $f(\{\alpha_i\},\{B_i\})$ is indeed a mixed graphon, (\emph{i.e.}, mixture of $k$ graphons $\{B_1, B_2, \cdots, B_k\}$), where
{each element $B_i$ represents a graphon estimated from a set of similar pre-training (sub)graphs.
Furthermore, it can be theoretically justified that the mixed version still preserve the properties of graphons (c.f. Theorem~\ref{theor-preserve}) and the key transferable patterns    inherited in $B_i$ (c.f. Theorem~\ref{theor:diff}).}
Thus the graph generator $f(\{\alpha_i\},\{B_i\})$, \emph{i.e.}, mixed graphon, serves as a representative and comprehensive summary of pre-training data,  from which unseen graphs with different combinations of transferable patterns can be induced.

\vpara{Possible downstream space.}
All the graphs produced by the generators in the generator space $\Omega$ could benefit from the pre-training, and finally form the possible downstream space.

Formally, for each generator in the generator space $\Omega$
(we denote it as $f$ for simplicity), we can generate a $n$-node graph as follows.
First, we independently sample a random latent variable for each node. Then for each pair of nodes, we assign an edge between them with the probability equal to the value of the graphon at their randomly sampled points.
This process can be formulated as:
\begin{equation}
\begin{aligned}
& v_1, \cdots, v_n \sim \textrm{Uniform}([0,1]), \\
& A_{ij} \sim \textrm{Bernouli}(
f(v_i,v_j))
, \quad \forall i,j \in \{1,2,...,n\},
\end{aligned}
\end{equation}
where $f(v_i,v_j) \in [0,1]$ indicates the corresponding value of the graphon  at point $(v_i ,v_j)$~\footnote{For simplicity, we slightly abuse the notations $f(\cdot,\cdot)$. Note that $f(\{\alpha_i\},\{B_i\})$ is a function of $\{\alpha_i\}$ and $\{B_i\}$, representing that the generator depends on $\{\alpha_i\}, \{B_i\}$; while for each generator (\emph{i.e.}, mixed graphon) $f$ given $\{\alpha_i\}, \{B_i\}$, it can be represented as a continuous, bounded and symmetric function $f: [0,1]^2 \rightarrow [0,1]$.}, and $A_{ij} \in \{0,1\}$ indicates the existence of edge between $i$-th node and $j$-th node.
The adjacency matrix of the sampled graph $G$ is denoted as $A = [A_{ij}] \in \{0,1\}^{n \times n}, \forall i,j \in[n]$.
We summarize this generation process as $G \shortleftarrow f$.

Therefore, with all generators from the generator space $\Omega$, 
the possible downstream space is defined as $\mathcal{D} = \{G \shortleftarrow f |f \in \Omega\}$.
Note that for each ${\{\alpha_i\},\{B_i\}}$,
we have a generator $f$; and for each generator, we also have different generated graphs.
Besides, we theoretically justify that the generated graphs in the possible downstream space can inherit key transferable graph patterns in our generator  (c.f. Theorem~\ref{thm:down}).

\vpara{Application cases.}
The proposed framework is flexible to be adopted in different application scenarios when discussing {the problem of} when to pre-train GNNs.

\begin{itemize}[leftmargin=*,topsep=0pt]
    \item \emph{Use case 1: provide a user guide of a graph pre-trained model.} The possible downstream space $\mathcal{D}$ serves as a user guide of a graph pre-trained model, telling the application scope  of graph pre-trained models (\emph{i.e.}, the possible downstream graphs that can benefit from the pre-training data).

 \item \emph{Use case 2:  estimate the feasibility of performing pre-training from pre-training data to downstream data.}
Given a collection of pre-training graphs and a downstream graph, one can directly measure the feasibility of performing pre-training on pre-training data, before conducting costly pre-training and fine-tuning attempts. By making such pre-judgement of a kind of transferability, some unnecessary and expensive parameter optimization steps during model training and evaluation can be avoided.

 \item \emph{Use case 3: select pre-training data to benefit the downstream.}
In some practical scenarios {where} the downstream data is provided (\emph{e.g.}, a company needs to boost downstream performance of its business data), the feasibility of pre-training inferred by W2PGNN can be used to select data for pre-training to maximize the downstream performance with limited resources.
\end{itemize}

Use case 1 can be directly given by our produced possible downstream space $\mathcal{D}$.
However, how to measure the feasibility of pre-training in use case 2 and 3 still remains a key challenge.
In the following sections, we introduce the formal definition of the feasibility of pre-training and its approximate solution.

\subsection{Feasibility of Pre-training}
~\label{subsec:fea}
If a downstream graph can be generated with a higher probability from any generator in the generator space $\Omega$, then the graph could benefit more from the pre-training data. 
We therefore define the feasibility of performing pre-training as the highest probability of the downstream data generated from a generator in $\Omega$, which can be formulated as an optimization problem as follows.

\begin{definition}[Feasibility of graph pre-training]\label{def:fea}
Given the pre-training data $\mathcal G_\text{train}$ and downstream data $\mathcal G_\text{down}$, we have the feasibility of performing pre-training on $\mathcal G_\text{train}$ to benefit $\mathcal G_\text{down}$ as
\begin{equation} \label{eq:fea}
    \zeta(\mathcal G_\text{train} 
    \shortrightarrow \mathcal G_\text{down}) = \underset{\{\alpha_i\},\{B_i\}}{\sup } \operatorname{Pr}\left(\mathcal G_\text{down} \mid f(\{\alpha_i\},\{B_i\})\right),
\end{equation}
where $\operatorname{Pr}\left(\mathcal G_\text{down} \mid f(\{\alpha_i\},\{B_i\}) \right)$ denotes the probability of 
the graph sequence sampled from $\mathcal G_\text{down}$ being generated by graph generator $f(\{\alpha_i\},\{B_i\})$; each (sub)graph represents an ego-network  (for node-level task) or a graph (for graph-level task) sampled from the downstream data $\mathcal{G}_\text{down}$.
\end{definition}

However, the probability $\operatorname{Pr}\left(\mathcal G_\text{down} \mid f(\{\alpha_i\},\{B_i\})\right)$ of generating the downstream graph from a generator is extremely hard to compute, we therefore {turn to converting} the optimization problem~\eqref{eq:fea} to a tractable problem.
Intuitively, if generator $f(\{\alpha_i\},\{B_i\})$ can generate the downstream data with higher probability, it potentially means that the underlying generative patterns of pre-training data (characterized by $f(\{\alpha_i\},\{B_i\})$) and downstream data (characterized by the graphon $B_\text{down}$ fitted from $\mathcal{G}_\text{down}$) are more similar.
Accordingly, we turn to figure out the infimum of the distance between $f(\{\alpha_i\},\{B_i\})$ and $B_\text{down}$ as the feasibility, \emph{i.e.},
\begin{equation}~\label{eq:dist}
  \zeta(\mathcal G_\text{train} 
    \shortrightarrow \mathcal G_\text{down}) =  - \underset{\{\alpha_i\},\{B_i\}}{\inf }  \operatorname{dist}(f(\{\alpha_i\},\{B_i\}), B_\text{down}).
\end{equation}
Following~\cite{Xu2021LearningGA}, we hire the 2-order Gromov-Wasserstein (GW) distance
as our distance function $\operatorname{dist}(\cdot,\cdot)$, as
GW distance is commonly used to measure the difference between structured data.

Additionally, we establish a theoretical connection between the above-mentioned distance and the probability of generating the downstream data in extreme case, which further adds to the integrity and rationality of our solution. Detailed proof of the following theorem can be found in Appendix~\ref{app:proof}.
\begin{theorem}\label{theor:dis-prob}
    Given the graph sequence sampled from downstream data $\mathcal{G}_\text{down}$, we estimate its corresponding graphon as $B_\text{down}$. If a generator $f$ can generate the downstream graph sequence with probability 1, then $\operatorname{dist}(f, B_\text{down})=0$.
\end{theorem}
% }

\subsection{Choose Graphon Basis to Approximate Feasibility}
\label{subsec:3.3}

Although the feasibility has been converted to the optimization problem~\eqref{eq:dist},
exhausting all possible $\{\alpha_i\},\{B_i\}$ to find the infimum is impractical. 
An intuitive idea is that we can choose some appropriate graphon {basis} $\{B_i\}$, which can not only prune the search space but also accelerate the optimization process.
Therefore, we aim to first reduce the search space of graphon basis $\{B_i\}$ and then learn the optimal $\{\alpha_i\}$ in the reduced search space.

Considering that the downstream data may be formed via {different generation mechanisms (implying various transferable patterns)}, a single graphon basis might have limited expressivity and completeness to cover all patterns. 
We therefore argue that a good reduced search space of graphon basis should cover a set of graphon bases. Here, we introduce three candidates of them as follows.

\vpara{Integrated graphon basis.}
The first candidate of graphon basis is the integrated graphon basis $\{B_i\}_\text{integr}$. This graphon basis is introduced based on the assumption that the pre-training and the downstream graphs share very similar patterns. For example, the pre-training and the downstream graphs might come from social networks of different time spans~\cite{Hu2020GPTGNNGP}. In the situation, almost all patterns involved in the pre-training data might be useful for the downstream.
To achieve this, we directly utilize all (sub)graphs sampled from the pre-training data to estimate one graphon as the graphon basis. This integrated graphon basis serves as a special case of the graphon basis introduced below.

\vpara{Domain graphon basis.}
The second candidate is the domain graphon basis $\{B_i\}_\text{domain}$. The domain information that pre-training data comes from is important prior knowledge to indicate the transferability from the pre-training to downstream data.
For example, when the downstream data is molecular network, it is more likely to benefit from the pre-training data from specific domains like biochemistry. This is because the specificity of molecules makes it difficult to learn transferable patterns from other domains, \emph{e.g.},
 closed triangle structure represents diametrically opposite meanings (stable vs unstable) in social network and molecular network.
Therefore, we propose to split the (sub)graphs sampled from  pre-training data according to their domains, and each split of (sub)graphs will be used to estimate a graphon as a basis element. In this way, each basis element reflects transferable patterns from a specific domain, and all basis elements construct the domain graphon basis $\{B_i\}_\text{domain}$.

\vpara{Topological graphon basis.}
The third candidate is the topological graphon basis  $\{B_i\}_\text{topo}$. The topological similarity between the pre-training and the downstream data serves as a crucial indicator of transferability. For example, a downstream social network might benefit from the similar topological patterns in academic or web networks (\emph{e.g.}, closed triangle structure indicates stable relationship in all these networks). 
Then, the problem of finding topological graphon basis can be converted to
partition $n$ (sub)graphs sampled from pre-training data into $k$-split according to their topology similarity, where each split contains (sub)graphs with similar topology. Each element of graphon basis (\emph{i.e.}, graphon) fitted from each split of (sub)graphs is expected to characterize a specific kind of topological transferable pattern.

However, the challenge {is} that for graph structured data that is irregular and complex, we cannot directly measure the topological similarity between graphs.
To tackle this problem, we introduce a \emph{graph feature extractor} that maps arbitrary graph into a fixed-length vector representation.
To approach a comprehensive and representative set of topological features,
we here consider both node-level and graph-level properties.

For node-level topological features, we first apply a set of node-level property functions $[\phi_1(v),\cdots,\phi_{m_1}(v)]$ for each node $v$ in graph $G$ to capture the local topological features around it. 
Considering that the numbers of nodes of two graphs are possibly different, we introduce an aggregation function $\operatorname{AGG}$ to summarize the node-level property of all nodes over $G$ to a real number $\operatorname{AGG}(\{{\phi}_i(v), v \in G\})$.
We can thus obtain the node-level topological vector representation as follows.
$$
h_\text{node} (G) =  [\operatorname{AGG}(\{{\phi}_1(v), v \in G\}), \cdots, \operatorname{AGG}(\{{\phi}_{m_1}(v), v \in G\})].
$$
In practice, we calculate degree~\cite{borner2007network}, clustering coefficient~\cite{Kaiser2008MeanCC} and closeness centrality~\cite{freeman2002centrality} for each node and instantiate the aggregation function $\operatorname{AGG}$ as the mean aggregator.

For graph-level topological features, we also employ a set of graph-level property functions for each graph $G$ to serve as the vector representation
$$
h_\text{graph}(G) = [\psi_1(G),\cdots,\psi_{m_2}(G)],
$$
where   density~\cite{wasserman1994social}, assortativity~\cite{newman2003mixing}, transitivity~\cite{wasserman1994social} are adopted as graph-level properties here.
~\footnote{Other graph-level properties can also be utilized like  \emph{diameter and Wiener index}, but we do not include them due to their high computational complexity.}.

Finally, the final representation of $G$ produced by the graph feature extractor is
$$
h=[h_\text{local}(G)||h_\text{global}(G)] \in \mathbb{R}^{m_1+m_2},
$$
where $||$  is the concatenation function that combines both node-level and graph-level features.
Given the topological vector representation, we leverage an efficient clustering algorithm  K-Means~\cite{macqueen1967classification} to obtain k-splits of (sub)graphs and finally fit each split into a graphon as one element of topological graphon basis.

\vpara{Optimization solution.}
Given the above-mentioned three graphon bases, the choice of graphon basis $\{B_i\}$ can be specified to one of them.
In this way, the pre-training feasibility (simplified as $\zeta$) could be approximated in the reduced search space of graphon basis as
\begin{equation}~\label{eq:dist3}
\begin{aligned}
\zeta \leftarrow -  \operatorname{MIN}  ( \{ \underset{\{\alpha_i\}}{\inf }  \operatorname{dist}(f(\{\alpha_i\},\{B_i\}), B_\text{down}),  \forall \{B_i\} \in \mathcal{B} \}),
\end{aligned}
\end{equation}
where $\mathcal{B}$$=$$\{\{B_i\}_\text{topo}, \{B_i\}_\text{domain}, \{B_i\}_\text{integr}\} $ is the reduced search space of $\{B_i\}$.
Thus, the problem can be naturally splitted into three sub-problems with objective of  $\operatorname{dist}(f(\{\alpha_i\},\{B_i\}_\text{topo}), B_\text{down}))$, $\operatorname{dist}\\(f(\{\alpha_i\}, \{B_i\}_\text{domain}), B_\text{down}))$ and $\operatorname{dist}(f(\{\alpha_i\},\{B_i\}_\text{integr}), B_\text{down}))$ respectively.
Each sub-problem can be solved by updating the corresponding learnable parameters $\{\alpha_i\}$ with multiple gradient descent steps. Taking one step as an example, we have
\begin{equation} \label{eq.2}
\begin{aligned}
    \{\alpha_i\} = \{\alpha_i\}-\eta \nabla_{\{\alpha_i\}} \operatorname{dist}(f(\{\alpha_i\},\{B_i\}), B_\text{down})
\end{aligned}
\end{equation}
where $\eta$ is the learning rate. 
Finally, we achieve three infimum distances under different $ \{B_i\} \in \mathcal B$ respectively, the minimum value among them is the approximation of pre-training feasibility.
In practice, we adopt an efficient and differential approximation of GW distance, \emph{i.e.}, entropic regularization GW distance~\cite{peyre2016gromov}, as the distance function.
{For graphon estimation, we use the ``largest gap'' method as to estimate graphon $B_i$.}

\subsection{Computation Complexity}~\label{subsec:cost}
We now show that the time complexity of W2PNN is much lower than traditional  solution.
Suppose that we have $n_1$  and $n_2$ (sub)graphs sampled from pre-training data and  downstream data respectively, and denote $|V|$ and $|E|$ as the average number of nodes and edges per (sub)graph.
The overall time complexity of W2PGNN is $O((n_1+n_2)|V|^2)$.
For comparison, traditional solution in Figure~\ref{fig:example}(a) to estimate the pre-training feasibility should make $l_1 \times l_2$ ``pre-train and fine-tune'' attempts, if there exist $l_1$ pre-training models and $l_2$ fine-tuning strategies. Suppose the batch size of pre-training as $b$ and the representation dimension as $d$. 
The overall time complexity of traditional solution is 
 $O\left(l_1l_2((n_1+n_2)(|V|^3 + |E|d)+n_1bd)\right)$.
Detailed analysis can be found in Appendix~\ref{app:complexity}.

\section{Theoretical Analysis}

In this section, we theoretically analyze the rationality of the generator space and possible downstream space in W2PGNN.  Detailed proofs of the following theorems can be found in Appendix~\ref{app:proof}.

\subsection{Theoretical Justification of Generator Space}
\vpara{Our generator preserves the properties of graphons.}
We first theoretically prove that any generator in the generator space still preserve the properties of graphon (\emph{i.e.}, a bounded symmetric function $ [0,1]^{2} \rightarrow [0,1]$, summarized in the following theorem.
\begin{theorem} \label{theor-preserve}
For a set of graphon basis $\{B_i\}$, the corresponding generator space $\Omega=\{ f(\{\alpha_i\},\{B_i\}) \mid \forall \ \{\alpha_i\},\{B_i\} \}$ is the convex hull of $\{B_i\}$.
\end{theorem}

\vpara{Our generator preserves the key transferable patterns in graphon basis.}
As a preliminary, we first introduce the concept of \emph{graph motifs} as a useful description of transferable graph patterns and leverage \emph{homomorphism density} as a measure to quantify the degree to which the patterns inherited in a graphon.
\begin{definition}[Graph motifs~\cite{milo2002network}]
Given a graph $G=(V,E)$ ($V$ and $E$ are node and edge set), graph motifs are substructures  $F=(V^\prime,E^\prime)$ that recur significantly in statistics, where $V^{\prime} \subset V, E^{\prime} \subset E$ and $\left|V^{\prime}\right| \ll|V|$.
\end{definition}
Graph motifs can be roughly taken as the key transferable graph patterns across graphs~\cite{zhang2021motif}. 
For example, the motif  ($\vcenter{\hbox{\includegraphics[width=2.4ex,height=2.4ex]{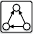}}}$) has the same meaning of ``feedforward loop'' across networks of control system, gene systems or organisms.

Then, we introduce the measure of homomorphism density  $t(F,B)$ to quantify the relative frequency of the key transferable pattern, \emph{i.e.}, graph motifs $F$, inherited in graphon $B$.
\begin{definition}[Homomorphism density~\cite{lovasz2012large}]
Consider a graph motif $F=(V^\prime,E^\prime)$, we define a homomorphisms of $F$ into graph $G=({V}, {E})$ as an adjacency-preserving map from $V^\prime$ to $V$, where 
$(i,j) \in {E}^\prime$ implies $(i,j) \in {E}$.  There could be multiple maps from $V^\prime$ to $V$, but only some of them are homomorphisms. Therefore, the definition of homomorphism density $t(F,G)$ is introduced to
quantify the relative frequency with which the graph motif $F$ appears in $G$.

Analogously, the homomorphism density of graphs can be extended into the graphon $B$.
We denote $t(F,B)$ as the homomorphism density of graph motif $F$ into graphon $B$, which represents the relative frequency of $F$ occurring in a collection of graphs  $\{{G}_i\}$ that convergent to graphon $B$, \emph{i.e.}, $ t({F}, B) =
\lim _{i \rightarrow \infty} t\left({F}, \{{G}_i\}\right)$.
\end{definition}

Now, we are ready to quantify how much the transferable patterns in graphon basis can be preserved in our generator by exploring the difference between the homomorphism density of graph motifs into the graphon basis and that into our generator.

\begin{theorem}\label{theor:diff}
Assume a graphon basis $\{B_1, \cdots, B_k\}$ and their convex combination $f( \{\alpha_i\}, \{B_i\}) = \sum _{i=1}^k \alpha_i B_i$. The $a$-th element of graphon basis $B_a$ corresponds to a motif set. For each motif $F_a$ in the motif set, the difference between the homomorphism density of $F_a$ in $f( \{\alpha_i\}, \{B_i\})$ and  that in  basis element $B_a$ is upper bounded by
\begin{equation}
|t(F_a,f(\{ \alpha_i\}, \{B_i\}))-t(F_a,B_a)|\leq \sum _{b=1,b\neq a}^k  |F_a| \alpha_b || B_b -B_a||_\square
\end{equation}
where $|F_a|$ represents the number of nodes in motif $F_a$,  $||\cdot||_\square$ is the cut norm.
\end{theorem}
Theorem~\ref{theor:diff} indicates the graph motifs (\emph{i.e.}, key transferable patterns) inherited in each basis element can be preserved in our generator, which justifies the rationality to take the generator as a representative and comprehensive summary of pre-training data.

\subsection{Theoretical Justification of Possible Downstream Space}

The possible downstream space includes the graphs generated from generator $f( \{\alpha_i\}, \{B_i\})$. We here provide a theoretical justification that the generated graphs in possible downstream space can inherit key transferable graph patterns (\emph{i.e.}, graph motifs) in the generator.

\begin{theorem}\label{thm:down}
Given a graph generator $f(\{ \alpha_i\}, \{B_i\})$, we can obtain sufficient number of random graphs $\mathbb{G} = \mathbb{G}(n, f( \{\alpha_i\}, \{B_i\}))$ with $n$ nodes 
generated from $f( \{\alpha_i\}, \{B_i\})$.
The homomorphism density of graph motif $F$ in $\mathbb{G}$ can be considered approximately equal to that in $f( \{\alpha_i\}, \{B_i\})$ with high probability and can be represented as
\begin{equation}
\mathrm{P}(|t(F, \mathbb{G})-t(F, f( \{\alpha_i\}, \{B_i\}))|>\varepsilon) \leq 2 \exp \left(-\frac{\varepsilon^{2} n}{8 \mathrm{v}(F)^{2}}\right),
\end{equation}
where $\mathrm{v}(F)$ denotes the number of nodes in $F$, and $0 \leq \epsilon \leq 1$. 
\end{theorem}
Theorem~\ref{thm:down} indicates that 
the homomorphism density of graph motifs into the generated graphs in the possible downstream space
can be inherited from our generator to a significant degree.

\section{Experiments}

In this section, we evaluate the effectiveness of W2PGNN
with the goal of answering the following questions: (1) Given the pre-training and downstream data, is the feasibility of pre-training estimated by W2PGNN positively correlated with the downstream performance (Use case 2)? 
(2) When the downstream data is provided, 
does the pre-training data selected by W2PGNN actually help improve the downstream performance (Use case 3)? 

Note that it is impractical to empirically evaluate the application scope of graph pre-trained models  (Use case 1), as we cannot enumerate all graphs in the possible downstream space. 
By answering question (1), it can be indirectly verified that a part of graphs in the possible downstream space, \emph{i.e.}, the downstream graphs with high feasibility, indeed benefit from the pre-training.

\begin{table*}[!t]
\centering
\resizebox{1.8\columnwidth}{!}{
\begin{tabular}{lcccccccccc}
\toprule[1pt]
\multicolumn{1}{c}{\multirow{2}{*}{}} & \multicolumn{5}{c}{$N=2$}                & \multicolumn{5}{c}{$N=3$}              \\
\cmidrule[1pt](r){2-6}\cmidrule[1pt](r){7-11} \multicolumn{1}{c}{}                         & US-Airport & Europe-Airport & H-index & Chameleon & Rank & US-Airport & Europe-Airport & H-index & Chameleon & Rank \\ \midrule[1pt]
Graph Statistics                             & -0.6068    & 0.3571         & -0.6220 & -0.2930   & 10   & -0.7096    & -0.5052        & -0.2930 & -0.8173   & 10  \\
EGI                                          & 0.0672     & -0.6077        & -0.2152 & -0.2680   & 9   & -0.2358    & -0.5540        & -0.2822 & -0.6511   & 9   \\
Clustering Coefficient                       & -0.0273    & 0.1519         & 0.3622  & 0.3130    & 5    & -0.0039    & 0.2069         & 0.4829  & 0.2279    & 4    \\
Spectrum of Graph Laplacian                  & -0.2023    & 0.1467         & 0.0794  & 0.0095    & 8    & -0.7648    & -0.4311        & 0.2611  & -0.2300   & 8    \\
Betweenness Centrality           & -0.2739    & -0.2554        & 0.2051  & 0.2241    & 7    & -0.3421    & -0.5903        & 0.1364  & 0.0849    & 7    \\ \midrule
W2PGNN (intergr)                             & 0.3579     & 0.1224         & 0.3313  & 0.1072    & 6    & 0.0841     & 0.5310         & 0.4213  & -0.0916   & 6    \\
W2PGNN (domain)                   & \textbf{0.4774}     & 0.4666         & 0.6775  & 0.3460    & 3    & \textbf{0.7132}     & 0.5523         & \textbf{0.7381}  & 0.1857    & 3    \\
W2PGNN (topo)                                & 0.2059     & 0.3908         & 0.3745  & 0.4464    & 4    & 0.4900     & 0.5061         & 0.4072  & 0.1497    & 5    \\
W2PGNN ($\alpha=1$)                          & 0.4172     & 0.5206         & 0.6829  & 0.4391    & 2    & 0.5282     & 0.6663         & 0.7240  & \textbf{0.3246}    & 1    \\
W2PGNN                                       & 0.3941     & \textbf{0.5336}         & \textbf{0.7162}  & \textbf{0.4838}    & 1    & 0.5089     & \textbf{0.6706}         & 0.6754  & 0.3166    & 2    \\ \bottomrule[1pt]
\end{tabular}
}
\caption{Pearson correlation coefficient between the estimated pre-training feasibility and the best downstream performance on node classification. {$N$ denotes the number of candidate pre-training datasets that form the pre-training data.} \textbf{Bold} indicates the highest coefficient. ``Rank'' represents the overall ranking on all downstream datasets.} 
 \vspace{-0.2in}
\label{tab:node_classification_results}
\end{table*} 

\begin{table*}[!t]
\centering
\resizebox{1.8\columnwidth}{!}{
\begin{tabular}{lcccccccccccc}
\toprule[1pt]
\multicolumn{1}{c}{\multirow{2}{*}{}} & \multicolumn{6}{c}{$N=2$}              & \multicolumn{6}{c}{$N=3$}            \\
\cmidrule[1pt](r){2-7}\cmidrule[1pt](r){8-13} \multicolumn{1}{c}{}                         & BACE    & BBBP    & MUV     & HIV     & ClinTox & Rank & BACE    & BBBP    & MUV     & HIV     & ClinTox & Rank \\ \midrule[1pt]
Graph Statistics                             & -0.4118 & -0.1328 & 0.3858  & 0.0174  & -0.3577 & 9    & -0.3093 & -0.1430 & 0.1946  & 0.3545  & -0.1372 & 7    \\
EGI                                          & 0.2912  & -0.6862 & 0.4488  & 0.0587  & 0.0452  & 7    & 0.4570  & 0.3230  & 0.3024  & 0.4144  & -0.0085 & 3    \\
Clustering Coefficient                       & -0.5098 & -0.5097 & 0.3754  & 0.4738  & 0.5154  & 8    & -0.4080 & 0.3217  & -0.1190 & -0.2483 & -0.4248 & 9   \\
Spectrum of Graph Laplacian                  & -0.0633 & -0.4878 & -0.3413 & -0.1125 & -0.2562 & 10   & -0.3563 & -0.1611 & -0.2294 & -0.2448 & 0.3001  & 8    \\
Betweenness Centrality           & -0.0021 & 0.7755  & 0.4040  & 0.0339  & 0.3411  & 6    & -0.3695 & -0.4568 & -0.2752 & -0.3035 & -0.2129 & 10   \\ \midrule
W2PGNN (intergr)                             & 0.7547  & \textbf{0.7790}  & 0.2907  & 0.7033  & 0.5639  & 3    & 0.4081  & 0.4687  & -0.0567 & 0.3802  & 0.4354  & 5    \\
W2PGNN (domain)                   & 0.7334  & 0.7689  & 0.5395  & 0.6831  & 0.5431  & 5    & 0.0864  & 0.3680  & 0.0187  & 0.4784  & 0.3765  & 6    \\
W2PGNN (topo)                                & 0.6656  & 0.7164  & {0.8131}  & \textbf{0.7391}  & 0.5406  & 2    & 0.1109  & 0.5357  & 0.0514  & 0.3265  & 0.4724  & 4    \\
W2PGNN ($\alpha=1$)                          & 0.6549  & 0.7690  & 0.6730  & 0.7033  & 0.5639  & 4    & 0.5287  & \textbf{0.7102}  & 0.1925  & 0.5893  & 0.5430  & 2    \\
W2PGNN                                       & \textbf{0.7549}  & 0.7776  & \textbf{0.8131}  & 0.7044  & \textbf{0.5784}  & 1    & \textbf{0.6207}  & 0.6696  & \textbf{0.5227}  & \textbf{0.6529}  & \textbf{0.5994}  & 1    \\ \bottomrule[1pt]
\end{tabular}
}
\caption{Pearson correlation coefficient between the  feasibility and the best downstream performance on graph classification.} 
 \vspace{-0.2in}
\label{tab:graph_classification_results}
\end{table*}

 \vspace{-0.1in}
\subsection{Experimental Setup}

We validate our proposed framework on both node classification and graph classification task. 

\vpara{Datasets.} 
For node classification task, we directly adopt six datasets from~\cite{Qiu2020GCCGC} as the candidates of pre-training data, which consists of Academia, DBLP(SNAP),
DBLP(NetRep), IMDB, Facebook and LiveJournal (from academic, movie and social domains). Regarding the downstream datasets, we adopt US-Airport and  H-Index from~\cite{Qiu2020GCCGC} and additionally add two more datasets Chameleon and Europe-Airport for a more comprehensive results.
For graph classification task, we choose the large-scale datasets ZINC15~\cite{Sterling2015ZINC1} containing 2 million unlabeled molecules.
To enrich the follow-up experimental analysis, we use scaffold split to partition the ZINC15 into five datasets ({ZINC15-0, ZINC15-1, ZINC15-2, ZINC15-3 and ZINC15-4}) according to their scaffolds~\cite{hu2019strategies}, such that the scaffolds are different in each dataset.
Regarding the downstream datasets, we use 5 classification benchmark
datasets BACE, BBBP, MUV, HIV and ClinTox  contained in MoleculeNet~\cite{wu2018moleculenet}. 
% The dataset details are summarized in Appendix~\ref{app:data}.

\vpara{Baseline of graph pre-training measures.} 
The baselines can be divided into 3 categories: 
(1) EGI~\cite{Zhu2021TransferLO} computes the difference between the graph Laplacian of (sub)graphs from pre-training data and downstream data;
(2) Graph Statistics, {by which} we merge average degree,  degree variance,  density,  degree assortativity coefficient,  transitivity and  average clustering coefficient to
construct a topological vector for each (sub)graph.
(3) Clustering Coefficient,
Spectrum of Graph Laplacian, and Betweenness Centrality, {by which} we adopt the distributions of graph properties as topological vectors.
For (2) and (3), we calculate the negative value of Maximum Mean Discrepancy distance between the obtained topological vectors of the (sub)graph from pre-training data and that from downstream data.
% For baselines, the distance/difference is computed between one ego-network (for node classification) or graph (for graph classification) from pre-training data and another one from downstream data.
For efficiency, when conducting node classification, we randomly sample 10\% nodes for each candidate pre-training dataset and all nodes for each downstream dataset, then extract their 2-hop ego-networks. 
The final measure is the average of distances/differences between each pair of pre-training and fine-tuning graphs.
% The final measure is the \JR{average?} of all distances/differences of each pair of pre-training and finetuning graphs.

\vpara{Implementation Details.} 
For node classification tasks, we randomly sample 1000 nodes for each pre-training dataset and extract 2-hops ego-networks of sampled nodes to compose our input space, and extract 2-hops ego-networks of all nodes in each downstream dataset to estimate the graphon.
For graph classification tasks, we take all graphs in each pre-training dataset to compose our input space and use all graphs in each downstream dataset to estimate graphon.
When constructing topological graphon basis, we set the the number of clusters $k=5$. The maximum iterations number of K-Means is set as $300$.  When constructing domain graphon basis, we take each pre-training dataset as a domain. 
For graphon estimation, we use the largest gap~\cite{channarond2012classification} approach and let {the block size of graphon } as the average number of nodes in all graphs.
When learning  ${\alpha_i}$,
we adopt Adam as the optimizer and set the learning rate $\eta$ as 0.05.
For the GW distance, we adopt its differential and efficient version entropic regularization GW distance with default hyperparameters~\cite{peyre2016gromov}. We
provide an open-source implementation of our model W2PGNN at https://github.com/caoyxuan/W2PGNN.
% \vspace{-0.2in}
\subsection{Results of Pre-training Feasibility}
\label{subsec:6.2}
\vpara{Setup.}
% When evaluating the pre-training feasibility, since its ground truth is unavailable,  we adopt the best downstream performance among a set of graph pre-training models as the ground truth.
{As a pre-judgement to assess the necessity of pre-training before conducting any pre-training/fine-tuning attempts, the graph pre-training feasibility should reveal the optimal case that downstream data can benefit from pre-training data. However, it is impractical to obtain the optimal case, because we cannot enumerate all factors affecting model performance, \emph{e.g.}, pre-training strategies, fine-tuning strategies, backbone models. Hence we use the best downstream performance achieved among existing commonly-used pre-training models as an approximation.}

For node classification tasks, we use the following 4 graph pre-training models: GraphCL~\cite{you2020graph} and GCC models~\cite{Qiu2020GCCGC} with three different  hyper-parameter (\emph{i.e.,}  128, 256 and 512 rw-hops).
For graph classification tasks, we adopt 7 SOTA
pre-training models:  AttrMasking~\cite{hu2019strategies}, ContextPred~\cite{hu2019strategies}, EdgePred~\cite{hu2019strategies},
Infomax~\cite{hu2019strategies}, GraphCL~\cite{you2020graph}, GraphMAE~\cite{hou2022graphmae} and JOAO~\cite{you2021graph}. 
When pre-training, we directly use the default hyper-parameters of pre-training models except the rw-hops in GCC.
During fine-tuning, we freeze the parameters of pre-trained models and utilize the logistic regression as classifier for node classification  and SVM as classifier for graph classification, following~\cite{Qiu2020GCCGC} and its fine-tuning  hyper-parameters. The downstream results are reported as the average of Micro F1 and  ROC-AUC under 10 runs on node classification and graph classification respectively. For each downstream task, we take the best performance among all methods. 

\begin{figure*}[!pt]
 \centering{\includegraphics[width=\linewidth]{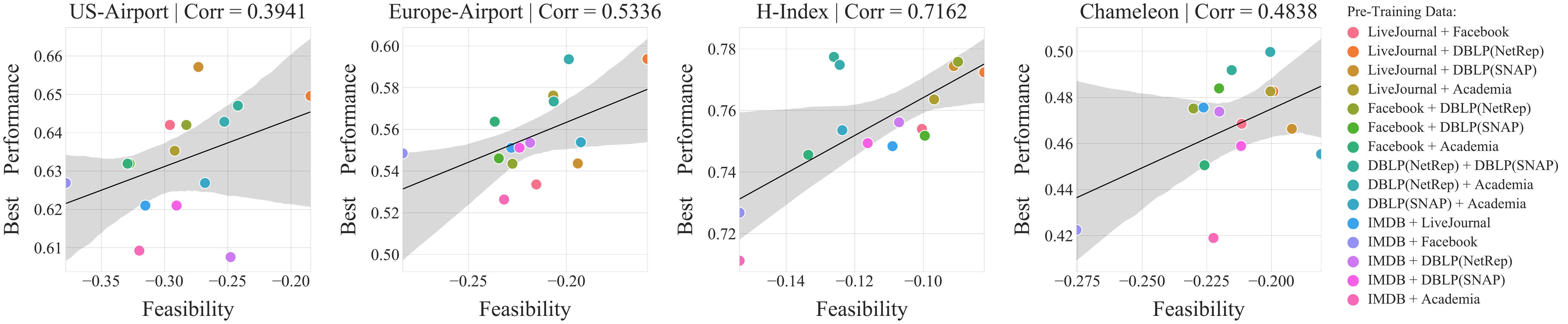}}
\caption{{Pre-training feasibility vs. the best downstream performance on node classification 
when the selection buget is 2.}  }
\label{fig:corre}
\vspace{-0.1in}
\end{figure*}

\begin{table*}[!h]
\centering
\resizebox{1.8\columnwidth}{!}{
\begin{tabular}{lcccccccccc}
\toprule[1pt]
\multicolumn{1}{c}{\multirow{2}{*}{}} & \multicolumn{5}{c}{$N=2$}                                                                                                                    & \multicolumn{5}{c}{$N=3$}     \\
\cmidrule[1pt](r){2-6}\cmidrule[1pt](r){7-11} \multicolumn{1}{c}{}                         & \multicolumn{1}{c}{US-Airport} & \multicolumn{1}{c}{Europe-Airport} & \multicolumn{1}{c}{H-index} & \multicolumn{1}{c}{Chameleon} & \multicolumn{1}{c}{Rank} & \multicolumn{1}{c}{US-Airport} & \multicolumn{1}{c}{Europe-Airport} & \multicolumn{1}{c}{H-index} & \multicolumn{1}{c}{Chameleon} & \multicolumn{1}{c}{Rank}  \\ \bottomrule[1pt]
\cellcolor{ggray!90} All Datasets                                      & \cellcolor{ggray!90} 65.62                          & \cellcolor{ggray!90} 55.65                              & \cellcolor{ggray!90}75.22                       & \cellcolor{ggray!90}46.81                         &   \cellcolor{ggray!90}-                     & \cellcolor{ggray!90}65.62                          & \cellcolor{ggray!90}55.65                              & \cellcolor{ggray!90}75.22                       & \cellcolor{ggray!90}46.81                                         &     \cellcolor{ggray!90} -                 \\
\toprule
Graph Statistics                             & 64.20                         &  53.36                            & 74.30                       & 44.31                         & 4                        & 62.27                          & 54.58                              & 72.88                        & 43.87                         & 5                           \\
EGI                                          &  \textbf{64.96}  &         57.37   &  74.30  &    43.21 & 2                        &  62.27    &       57.36 &   72.88 &45.93                        & 3                          \\
Clustering Coefficient                       &  62.61  &         52.87 &   \textbf{77.74}     & 43.21   & 3                        & 62.94      &     54.58 &   75.18     & 44.66 & 4                      \\
Spectrum of Graph Laplacian                  &  61.76 &          \textbf{57.88} &    73.14   &   42.20  & 5                        & \textbf{63.95}        &   54.87  &  73.90    &  44.66    & 2                         \\
Betweenness Centrality           &  \textbf{64.96} &          52.87 &   73.50   &   41.63  & 6                      & 62.27         &  54.87&  75.18 &     43.87    & 6                       \\ \midrule
W2PGNN                                       & \textbf{64.96}                       & \textbf{57.88}                             & 77.24               & \textbf{45.54}                         & 1                        & \textbf{63.95}                         & \textbf{57.59}                            & \textbf{75.68}                      & \textbf{46.07}                         & 1  \\  \bottomrule[1pt]  
\end{tabular}
}
\caption{Node classification results when performing pre-training on different selected pre-training data. We also provide the results of using all pre-training data without selection for your reference (see ``All Datasets'' in the table).}
\label{tab:node_classification_data_selection_results}
 \vspace{-0.2in}
\end{table*}

For a comprehensive evaluation on the correlation between the estimated pre-training feasibility and the best downstream performance, we need to construct multiple $\langle \mathcal G_\text{train}, \mathcal G_\text{down} \rangle$ sample pairs as our evaluation samples.
When constructing the $\langle \mathcal G_\text{train}, \mathcal G_\text{down} \rangle$ sample pairs for each downstream data, multiple pre-training data are required to be paired with it.
Hence we adopt the following two settings to augment the choice of pre-training data for more possibilities. 
We use $N$ as the number of dataset candidates contained in pre-training data.
% (1) For $N=2$, we randomly select 2 pre-training dataset candidates as pre-training data and enumerate all possible cases.
% (2) For $N=3$, we randomly select 3 pre-training dataset candidates as pre-training data. 
{For $N=2$ and $N=3$, we randomly select 2 and 3 pre-training dataset candidates, respectively as pre-training data.}
We enumerate all possible combination cases for graph classification tasks and randomly select 40\% of all cases for node classification tasks for efficiency.

\vpara{Results.} 
Table~\ref{tab:node_classification_results} (for node classification) and Table~\ref{tab:graph_classification_results} (for graph classification) show the Pearson correlation coefficient between the best downstream performance and the estimated pre-training feasibility by W2PGNN and baselines for each downstream dataset.  A higher coefficient indicates a better estimation of pre-training feasibility.
We also include 4 variants of W2PGNN:  W2PGNN (intergr), W2PGNN (domain) and W2PGNN (topo) only utilize the integrated graphon basis, domain graphon basis and topological graphon basis to approximate feasibility respectively, and W2PGNN ($\alpha=1$) set the learnable  combination weights $\{\alpha_i\}$ as constant 1.
We have the following observations.
(1) The results show that our model achieve the highest overall ranking in most cases, indicating the superiority of our proposed framework. 
(2) We find that the measures provided by other baselines sometimes show no correlation or negative correlation with the best downstream performance. 
(3) Comparing W2PGNN and its variants, we find that although the variants sometimes achieve superior performance on some downstream datasets, they cannot consistently perform well on all datastes. In contrast, the top-ranked W2PGNN can provide a more comprehensive picture with various graph bases and learnable combination weights.

To provide a deeper understanding of the feasibility estimated by W2PGNN, Figure~\ref{fig:corre} shows our estimated pre-training feasibility (in x-axis) versus the best downstream performance on node classification (in y-axis) of all <pre-training data, downstream data> pairs (one point represents the result of one pair) when the selection budget is 2.
The plots when the selection budget is 3 and the plots under graph classification can be found in Appendix~\ref{app:graph_class_res}.
We find that there is a strong positive correlation between estimated pre-training feasibility and the best downstream performance on all downstream datasets, which also suggests the significance of our feasibility.

 \vspace{-0.15in}
\subsection{Results of Pre-Training Data Selection}
Given the downstream data,  a collection of pre-training dataset candidates and a selection budget (\emph{i.e.}, the number of datasets selected for pre-training) due to limited resources, we aim to select the pre-training data with the highest feasibility, so as to benefit the downstream performance.

\vpara{Setup.} 
We here adopt two settings, \emph{i.e.}, selection budget is set as 2 and 3 respectively. The datasets that are augmented for more pre-training data choices in Section~\ref{subsec:6.2} can be directly used as the candidates of pre-training datasets here.
Then, the selected pre-training data serves as the input of graph pre-training model.
For node classification tasks, we adopt GCC as the pre-training model as an example, because it is the   pre-training model that can be generalized across domains and most of the datasets used for node classification are taken from it~\cite{Qiu2020GCCGC}.
For graph classification tasks, we take GraphCL as the pre-training model as it provides multiple graph augmentation approaches and is more general~\cite{you2020graph}.

\vpara{Results.}
Table~\ref{tab:node_classification_data_selection_results} shows the results of pre-training data selection on node classification task. (The results on graph classification is included in Appendix~\ref{app:graph_class_res2}). We have the following observations.
(1) We can see that the pre-training data selected by W2PGNN ranks first, which is the most suitable one for downstream.
(2) We find that sometimes simple graph property like clustering coefficient serves as a good choice on a specific dataset (\emph{i.e.,} H-index), when the budget of pre-training data is 2. It is because that H-index exhibits the largest clustering coefficient compared to other downstream datasets,  which facilitates the data selection via clustering coefficient.
However, such simple graph property is only applicable when the downstream dataset shows a strong indicator of the property,
and is not helpful when you need to select more datasets for pre-training (see results under $N$=3).
(3) Moreover, it is also interesting to see that using all pre-training data for pre-training is not always a {reliable} choice. We find that carefully selecting pre-training data can not only benefit downstream performance but also reduce computation resources.

\vspace{-0.1in}
\section{Conclusion}
This paper proposes a W2PGNN framework to answer the question of \emph{when to pre-train} GNNs based on the generative mechanisms from pre-training to downstream data. 
W2PGNN designs a graphon-based graph generator to summarize the knowledge in pre-training data, and the generator can in turn produce the solution space of downstream data that can benefit from the pre-training.
W2PGNN is theoretically and empirically shown to have great potential to provide the application scope of graph pre-training models, estimate the feasibility of pre-training and help select pre-training data.

\begin{acks}
This work was partially supported by NSFC (62206056), Zhejiang NSF (LR22F020005), the National Key Research and Development Project of China (2018AAA0101900), and the Fundamental Research Funds for the Central Universities.
\end{acks}
\bibliographystyle{ACM-Reference-Format}
\balance
\bibliography{reference}


\begin{thebibliography}{60}

%%% the Web address.

\ifx \showCODEN    \undefined \def \showCODEN     #1{\unskip}     \fi
\ifx \showDOI      \undefined \def \showDOI       #1{#1}\fi
\ifx \showISBNx    \undefined \def \showISBNx     #1{\unskip}     \fi
\ifx \showISBNxiii \undefined \def \showISBNxiii  #1{\unskip}     \fi
\ifx \showISSN     \undefined \def \showISSN      #1{\unskip}     \fi
\ifx \showLCCN     \undefined \def \showLCCN      #1{\unskip}     \fi
\ifx \shownote     \undefined \def \shownote      #1{#1}          \fi
\ifx \showarticletitle \undefined \def \showarticletitle #1{#1}   \fi
\ifx \showURL      \undefined \def \showURL       {\relax}        \fi
% invisible to TeX
\providecommand\bibfield[2]{#2}
\providecommand\bibinfo[2]{#2}
\providecommand\natexlab[1]{#1}
\providecommand\showeprint[2][]{arXiv:#2}

\bibitem[Airoldi et~al\mbox{.}(2013)]%
        {Airoldi2013StochasticBA}
\bibfield{author}{\bibinfo{person}{Edo~M Airoldi}, \bibinfo{person}{Thiago~B
  Costa}, {and} \bibinfo{person}{Stanley~H Chan}.}
  \bibinfo{year}{2013}\natexlab{}.
\newblock \showarticletitle{Stochastic blockmodel approximation of a graphon:
  Theory and consistent estimation}. In \bibinfo{booktitle}{\emph{NeurIPS}}.
  \bibinfo{pages}{692--700}.
\newblock


\bibitem[Backstrom et~al\mbox{.}(2006)]%
        {Backstrom2006GroupFI}
\bibfield{author}{\bibinfo{person}{Lars Backstrom}, \bibinfo{person}{Daniel~P.
  Huttenlocher}, \bibinfo{person}{Jon~M. Kleinberg}, {and}
  \bibinfo{person}{Xiangyang Lan}.} \bibinfo{year}{2006}\natexlab{}.
\newblock \showarticletitle{Group formation in large social networks:
  membership, growth, and evolution}. In \bibinfo{booktitle}{\emph{SIGKDD}}.
\newblock


\bibitem[B{\"o}rner et~al\mbox{.}(2007)]%
        {borner2007network}
\bibfield{author}{\bibinfo{person}{Katy B{\"o}rner}, \bibinfo{person}{Soma
  Sanyal}, \bibinfo{person}{Alessandro Vespignani}, {et~al\mbox{.}}}
  \bibinfo{year}{2007}\natexlab{}.
\newblock \showarticletitle{Network science}.
\newblock \bibinfo{journal}{\emph{Annu. rev. inf. sci. technol.}}
  \bibinfo{volume}{41}, \bibinfo{number}{1} (\bibinfo{year}{2007}),
  \bibinfo{pages}{537--607}.
\newblock


\bibitem[Channarond et~al\mbox{.}(2012)]%
        {channarond2012classification}
\bibfield{author}{\bibinfo{person}{Antoine Channarond},
  \bibinfo{person}{Jean-Jacques Daudin}, {and} \bibinfo{person}{St{\'e}phane
  Robin}.} \bibinfo{year}{2012}\natexlab{}.
\newblock \showarticletitle{Classification and estimation in the Stochastic
  Blockmodel based on the empirical degrees}.
\newblock \bibinfo{journal}{\emph{Electronic Journal of Statistics}}
  \bibinfo{volume}{6} (\bibinfo{year}{2012}), \bibinfo{pages}{2574--2601}.
\newblock


\bibitem[Devlin et~al\mbox{.}(2019)]%
        {devlin2018bert}
\bibfield{author}{\bibinfo{person}{Jacob Devlin}, \bibinfo{person}{Ming-Wei
  Chang}, \bibinfo{person}{Kenton Lee}, {and} \bibinfo{person}{Kristina
  Toutanova}.} \bibinfo{year}{2019}\natexlab{}.
\newblock \showarticletitle{Bert: Pre-training of deep bidirectional
  transformers for language understanding}. In
  \bibinfo{booktitle}{\emph{NAACL-HLT}}. \bibinfo{pages}{4171--4186}.
\newblock


\bibitem[Donnat et~al\mbox{.}(2018)]%
        {Donnat2018LearningSN}
\bibfield{author}{\bibinfo{person}{Claire Donnat}, \bibinfo{person}{Marinka
  Zitnik}, \bibinfo{person}{David Hallac}, {and} \bibinfo{person}{Jure
  Leskovec}.} \bibinfo{year}{2018}\natexlab{}.
\newblock \showarticletitle{Learning Structural Node Embeddings via Diffusion
  Wavelets}. In \bibinfo{booktitle}{\emph{SIGKDD}}.
\newblock


\bibitem[Freeman et~al\mbox{.}(2002)]%
        {freeman2002centrality}
\bibfield{author}{\bibinfo{person}{Linton~C Freeman} {et~al\mbox{.}}}
  \bibinfo{year}{2002}\natexlab{}.
\newblock \showarticletitle{Centrality in social networks: Conceptual
  clarification}.
\newblock \bibinfo{journal}{\emph{Social network: critical concepts in
  sociology. Londres: Routledge}}  \bibinfo{volume}{1} (\bibinfo{year}{2002}),
  \bibinfo{pages}{238--263}.
\newblock


\bibitem[Grover and Leskovec(2016)]%
        {Grover2016node2vecSF}
\bibfield{author}{\bibinfo{person}{Aditya Grover} {and} \bibinfo{person}{Jure
  Leskovec}.} \bibinfo{year}{2016}\natexlab{}.
\newblock \showarticletitle{node2vec: Scalable Feature Learning for Networks}.
  In \bibinfo{booktitle}{\emph{SIGKDD}}.
\newblock


\bibitem[Hafidi et~al\mbox{.}(2020)]%
        {Hafidi2020GraphCLCS}
\bibfield{author}{\bibinfo{person}{Hakim Hafidi}, \bibinfo{person}{Mounir
  Ghogho}, \bibinfo{person}{Philippe Ciblat}, {and} \bibinfo{person}{Ananthram
  Swami}.} \bibinfo{year}{2020}\natexlab{}.
\newblock \showarticletitle{GraphCL: Contrastive Self-Supervised Learning of
  Graph Representations}.
\newblock \bibinfo{journal}{\emph{ArXiv}}  \bibinfo{volume}{abs/2007.08025}
  (\bibinfo{year}{2020}).
\newblock


\bibitem[Han et~al\mbox{.}(2021)]%
        {Han2021AdaptiveTL}
\bibfield{author}{\bibinfo{person}{Xueting Han}, \bibinfo{person}{Zhenhuan
  Huang}, \bibinfo{person}{Bang An}, {and} \bibinfo{person}{Jing Bai}.}
  \bibinfo{year}{2021}\natexlab{}.
\newblock \showarticletitle{Adaptive Transfer Learning on Graph Neural
  Networks}. In \bibinfo{booktitle}{\emph{SIGKDD}}.
\newblock


\bibitem[Han et~al\mbox{.}(2022)]%
        {Han2022GMixupGD}
\bibfield{author}{\bibinfo{person}{Xiaotian Han}, \bibinfo{person}{Zhimeng
  Jiang}, \bibinfo{person}{Ninghao Liu}, {and} \bibinfo{person}{Xia Hu}.}
  \bibinfo{year}{2022}\natexlab{}.
\newblock \showarticletitle{G-Mixup: Graph Data Augmentation for Graph
  Classification}. In \bibinfo{booktitle}{\emph{ICML}}.
\newblock


\bibitem[Hassani and Khasahmadi(2020)]%
        {hassani2020contrastive}
\bibfield{author}{\bibinfo{person}{Kaveh Hassani} {and}
  \bibinfo{person}{Amir~Hosein Khasahmadi}.} \bibinfo{year}{2020}\natexlab{}.
\newblock \showarticletitle{Contrastive multi-view representation learning on
  graphs}. In \bibinfo{booktitle}{\emph{ICML}}. PMLR,
  \bibinfo{pages}{4116--4126}.
\newblock


\bibitem[He et~al\mbox{.}(2020)]%
        {he2020momentum}
\bibfield{author}{\bibinfo{person}{Kaiming He}, \bibinfo{person}{Haoqi Fan},
  \bibinfo{person}{Yuxin Wu}, \bibinfo{person}{Saining Xie}, {and}
  \bibinfo{person}{Ross Girshick}.} \bibinfo{year}{2020}\natexlab{}.
\newblock \showarticletitle{Momentum contrast for unsupervised visual
  representation learning}. In \bibinfo{booktitle}{\emph{CVPR}}.
  \bibinfo{pages}{9729--9738}.
\newblock


\bibitem[Hou et~al\mbox{.}(2022)]%
        {hou2022graphmae}
\bibfield{author}{\bibinfo{person}{Zhenyu Hou}, \bibinfo{person}{Xiao Liu},
  \bibinfo{person}{Yukuo Cen}, \bibinfo{person}{Yuxiao Dong},
  \bibinfo{person}{Hongxia Yang}, \bibinfo{person}{Chunjie Wang}, {and}
  \bibinfo{person}{Jie Tang}.} \bibinfo{year}{2022}\natexlab{}.
\newblock \showarticletitle{GraphMAE: Self-Supervised Masked Graph
  Autoencoders}. In \bibinfo{booktitle}{\emph{SIGKDD}}.
  \bibinfo{pages}{594--604}.
\newblock


\bibitem[Hu et~al\mbox{.}(2020b)]%
        {hu2019strategies}
\bibfield{author}{\bibinfo{person}{Weihua Hu}, \bibinfo{person}{Bowen Liu},
  \bibinfo{person}{Joseph Gomes}, \bibinfo{person}{Marinka Zitnik},
  \bibinfo{person}{Percy Liang}, \bibinfo{person}{Vijay Pande}, {and}
  \bibinfo{person}{Jure Leskovec}.} \bibinfo{year}{2020}\natexlab{b}.
\newblock \showarticletitle{Strategies for pre-training graph neural networks}.
  In \bibinfo{booktitle}{\emph{ICLR}}.
\newblock


\bibitem[Hu et~al\mbox{.}(2020a)]%
        {Hu2020GPTGNNGP}
\bibfield{author}{\bibinfo{person}{Ziniu Hu}, \bibinfo{person}{Yuxiao Dong},
  \bibinfo{person}{Kuansan Wang}, \bibinfo{person}{Kai-Wei Chang}, {and}
  \bibinfo{person}{Yizhou Sun}.} \bibinfo{year}{2020}\natexlab{a}.
\newblock \showarticletitle{GPT-GNN: Generative Pre-Training of Graph Neural
  Networks}. In \bibinfo{booktitle}{\emph{SIGKDD}}.
\newblock


\bibitem[Hu et~al\mbox{.}(2019)]%
        {Hu2019PreTrainingGN}
\bibfield{author}{\bibinfo{person}{Ziniu Hu}, \bibinfo{person}{Changjun Fan},
  \bibinfo{person}{Ting Chen}, \bibinfo{person}{Kai-Wei Chang}, {and}
  \bibinfo{person}{Yizhou Sun}.} \bibinfo{year}{2019}\natexlab{}.
\newblock \showarticletitle{Pre-Training Graph Neural Networks for Generic
  Structural Feature Extraction}.
\newblock \bibinfo{journal}{\emph{ArXiv}}  \bibinfo{volume}{abs/1905.13728}
  (\bibinfo{year}{2019}).
\newblock


\bibitem[Kaiser(2008)]%
        {Kaiser2008MeanCC}
\bibfield{author}{\bibinfo{person}{Marcus Kaiser}.}
  \bibinfo{year}{2008}\natexlab{}.
\newblock \showarticletitle{Mean clustering coefficients: the role of isolated
  nodes and leafs on clustering measures for small-world networks}.
\newblock \bibinfo{journal}{\emph{New Journal of Physics}}
  \bibinfo{volume}{10} (\bibinfo{year}{2008}), \bibinfo{pages}{083042}.
\newblock


\bibitem[Kipf and Welling(2016)]%
        {Kipf2016VariationalGA}
\bibfield{author}{\bibinfo{person}{Thomas Kipf} {and} \bibinfo{person}{Max
  Welling}.} \bibinfo{year}{2016}\natexlab{}.
\newblock \showarticletitle{Variational Graph Auto-Encoders}.
\newblock \bibinfo{journal}{\emph{ArXiv}}  \bibinfo{volume}{abs/1611.07308}
  (\bibinfo{year}{2016}).
\newblock


\bibitem[Li et~al\mbox{.}(2021)]%
        {Li2021PairwiseHD}
\bibfield{author}{\bibinfo{person}{Pengyong Li}, \bibinfo{person}{Jun Wang},
  \bibinfo{person}{Ziliang Li}, \bibinfo{person}{Yixuan Qiao},
  \bibinfo{person}{Xianggen Liu}, \bibinfo{person}{Fei Ma},
  \bibinfo{person}{Peng Gao}, \bibinfo{person}{Sen Song}, {and}
  \bibinfo{person}{Guowang Xie}.} \bibinfo{year}{2021}\natexlab{}.
\newblock \showarticletitle{Pairwise Half-graph Discrimination: A Simple
  Graph-level Self-supervised Strategy for Pre-training Graph Neural Networks}.
  In \bibinfo{booktitle}{\emph{IJCAI}}.
\newblock


\bibitem[Li et~al\mbox{.}(2022)]%
        {li2022let}
\bibfield{author}{\bibinfo{person}{Sihang Li}, \bibinfo{person}{Xiang Wang},
  \bibinfo{person}{An Zhang}, \bibinfo{person}{Yingxin Wu},
  \bibinfo{person}{Xiangnan He}, {and} \bibinfo{person}{Tat-Seng Chua}.}
  \bibinfo{year}{2022}\natexlab{}.
\newblock \showarticletitle{Let Invariant Rationale Discovery Inspire Graph
  Contrastive Learning}. In \bibinfo{booktitle}{\emph{ICML}}.
  \bibinfo{pages}{13052--13065}.
\newblock


\bibitem[Liu et~al\mbox{.}(2022)]%
        {liu2022user}
\bibfield{author}{\bibinfo{person}{Can Liu}, \bibinfo{person}{Yuncong Gao},
  \bibinfo{person}{Li Sun}, \bibinfo{person}{Jinghua Feng},
  \bibinfo{person}{Hao Yang}, {and} \bibinfo{person}{Xiang Ao}.}
  \bibinfo{year}{2022}\natexlab{}.
\newblock \showarticletitle{User Behavior Pre-training for Online Fraud
  Detection}. In \bibinfo{booktitle}{\emph{Proceedings of the 28th ACM SIGKDD
  Conference on Knowledge Discovery and Data Mining}}.
  \bibinfo{pages}{3357--3365}.
\newblock


\bibitem[Lov{\'a}sz(2012)]%
        {lovasz2012large}
\bibfield{author}{\bibinfo{person}{L{\'a}szl{\'o} Lov{\'a}sz}.}
  \bibinfo{year}{2012}\natexlab{}.
\newblock \bibinfo{booktitle}{\emph{Large networks and graph limits}}.
  Vol.~\bibinfo{volume}{60}.
\newblock \bibinfo{publisher}{American Mathematical Soc.}
\newblock


\bibitem[Lov{\'a}sz and Szegedy(2006)]%
        {lovasz2006limits}
\bibfield{author}{\bibinfo{person}{L{\'a}szl{\'o} Lov{\'a}sz} {and}
  \bibinfo{person}{Bal{\'a}zs Szegedy}.} \bibinfo{year}{2006}\natexlab{}.
\newblock \showarticletitle{Limits of dense graph sequences}.
\newblock \bibinfo{journal}{\emph{Journal of Combinatorial Theory, Series B}}
  \bibinfo{volume}{96}, \bibinfo{number}{6} (\bibinfo{year}{2006}),
  \bibinfo{pages}{933--957}.
\newblock


\bibitem[Lu et~al\mbox{.}(2021)]%
        {Lu2021LearningTP}
\bibfield{author}{\bibinfo{person}{Yuanfu Lu}, \bibinfo{person}{Xunqiang
  Jiang}, \bibinfo{person}{Yuan Fang}, {and} \bibinfo{person}{Chuan Shi}.}
  \bibinfo{year}{2021}\natexlab{}.
\newblock \showarticletitle{Learning to Pre-train Graph Neural Networks}. In
  \bibinfo{booktitle}{\emph{AAAI}}.
\newblock


\bibitem[MacQueen(1967)]%
        {macqueen1967classification}
\bibfield{author}{\bibinfo{person}{J MacQueen}.}
  \bibinfo{year}{1967}\natexlab{}.
\newblock \showarticletitle{Classification and analysis of multivariate
  observations}. In \bibinfo{booktitle}{\emph{5th Berkeley Symp. Math. Statist.
  Probability}}. University of California Los Angeles LA USA,
  \bibinfo{pages}{281--297}.
\newblock


\bibitem[Milo et~al\mbox{.}(2002)]%
        {milo2002network}
\bibfield{author}{\bibinfo{person}{Ron Milo}, \bibinfo{person}{Shai Shen-Orr},
  \bibinfo{person}{Shalev Itzkovitz}, \bibinfo{person}{Nadav Kashtan},
  \bibinfo{person}{Dmitri Chklovskii}, {and} \bibinfo{person}{Uri Alon}.}
  \bibinfo{year}{2002}\natexlab{}.
\newblock \showarticletitle{Network motifs: simple building blocks of complex
  networks}.
\newblock \bibinfo{journal}{\emph{Science}} \bibinfo{volume}{298},
  \bibinfo{number}{5594} (\bibinfo{year}{2002}), \bibinfo{pages}{824--827}.
\newblock


\bibitem[Narayanan et~al\mbox{.}(2017)]%
        {Narayanan2017graph2vecLD}
\bibfield{author}{\bibinfo{person}{Annamalai Narayanan},
  \bibinfo{person}{Mahinthan Chandramohan}, \bibinfo{person}{Rajasekar
  Venkatesan}, \bibinfo{person}{Lihui Chen}, \bibinfo{person}{Yang Liu}, {and}
  \bibinfo{person}{Shantanu Jaiswal}.} \bibinfo{year}{2017}\natexlab{}.
\newblock \showarticletitle{graph2vec: Learning Distributed Representations of
  Graphs}.
\newblock \bibinfo{journal}{\emph{ArXiv}}  \bibinfo{volume}{abs/1707.05005}.
\newblock


\bibitem[Newman(2003)]%
        {newman2003mixing}
\bibfield{author}{\bibinfo{person}{Mark~EJ Newman}.}
  \bibinfo{year}{2003}\natexlab{}.
\newblock \showarticletitle{Mixing patterns in networks}.
\newblock \bibinfo{journal}{\emph{Physical review E}} \bibinfo{volume}{67},
  \bibinfo{number}{2} (\bibinfo{year}{2003}), \bibinfo{pages}{026126}.
\newblock


\bibitem[Perozzi et~al\mbox{.}(2014)]%
        {Perozzi2014DeepWalkOL}
\bibfield{author}{\bibinfo{person}{Bryan Perozzi}, \bibinfo{person}{Rami
  Al-Rfou}, {and} \bibinfo{person}{Steven Skiena}.}
  \bibinfo{year}{2014}\natexlab{}.
\newblock \showarticletitle{DeepWalk: online learning of social
  representations}. In \bibinfo{booktitle}{\emph{SIGKDD}}.
\newblock


\bibitem[Peyr{\'e} et~al\mbox{.}(2016)]%
        {peyre2016gromov}
\bibfield{author}{\bibinfo{person}{Gabriel Peyr{\'e}}, \bibinfo{person}{Marco
  Cuturi}, {and} \bibinfo{person}{Justin Solomon}.}
  \bibinfo{year}{2016}\natexlab{}.
\newblock \showarticletitle{Gromov-wasserstein averaging of kernel and distance
  matrices}. In \bibinfo{booktitle}{\emph{ICML}}. PMLR,
  \bibinfo{pages}{2664--2672}.
\newblock


\bibitem[Qiu et~al\mbox{.}(2020)]%
        {Qiu2020GCCGC}
\bibfield{author}{\bibinfo{person}{Jiezhong Qiu}, \bibinfo{person}{Qibin Chen},
  \bibinfo{person}{Yuxiao Dong}, \bibinfo{person}{Jing Zhang},
  \bibinfo{person}{Hongxia Yang}, \bibinfo{person}{Ming Ding},
  \bibinfo{person}{Kuansan Wang}, {and} \bibinfo{person}{Jie Tang}.}
  \bibinfo{year}{2020}\natexlab{}.
\newblock \showarticletitle{GCC: Graph Contrastive Coding for Graph Neural
  Network Pre-Training}. In \bibinfo{booktitle}{\emph{SIGKDD}}.
\newblock


\bibitem[Ribeiro et~al\mbox{.}(2017)]%
        {Ribeiro2017struc2vecLN}
\bibfield{author}{\bibinfo{person}{Leonardo F.~R. Ribeiro},
  \bibinfo{person}{Pedro H.~P. Saverese}, {and} \bibinfo{person}{Daniel~R.
  Figueiredo}.} \bibinfo{year}{2017}\natexlab{}.
\newblock \showarticletitle{struc2vec: Learning Node Representations from
  Structural Identity}.
\newblock \bibinfo{journal}{\emph{SIGKDD}} (\bibinfo{year}{2017}).
\newblock


\bibitem[Ritchie et~al\mbox{.}(2016)]%
        {Ritchie2016ASP}
\bibfield{author}{\bibinfo{person}{Scott~C. Ritchie},
  \bibinfo{person}{Stephen~C. Watts}, \bibinfo{person}{Liam~G. Fearnley},
  \bibinfo{person}{Kathryn~E. Holt}, \bibinfo{person}{Gad Abraham}, {and}
  \bibinfo{person}{Michael Inouye}.} \bibinfo{year}{2016}\natexlab{}.
\newblock \showarticletitle{A Scalable Permutation Approach Reveals Replication
  and Preservation Patterns of Network Modules in Large Datasets.}
\newblock \bibinfo{journal}{\emph{Cell systems}}  \bibinfo{volume}{3 1}
  (\bibinfo{year}{2016}), \bibinfo{pages}{71--82}.
\newblock


\bibitem[Rozemberczki et~al\mbox{.}(2021)]%
        {Rozemberczki2021MultiscaleAN}
\bibfield{author}{\bibinfo{person}{Benedek Rozemberczki}, \bibinfo{person}{Carl
  Allen}, {and} \bibinfo{person}{Rik Sarkar}.} \bibinfo{year}{2021}\natexlab{}.
\newblock \showarticletitle{Multi-scale Attributed Node Embedding}.
\newblock \bibinfo{journal}{\emph{J. Complex Networks}}  \bibinfo{volume}{9}
  (\bibinfo{year}{2021}).
\newblock


\bibitem[Sterling and Irwin(2015)]%
        {Sterling2015ZINC1}
\bibfield{author}{\bibinfo{person}{T. Sterling} {and} \bibinfo{person}{John~J.
  Irwin}.} \bibinfo{year}{2015}\natexlab{}.
\newblock \showarticletitle{ZINC 15 – Ligand Discovery for Everyone}.
\newblock \bibinfo{journal}{\emph{Journal of Chemical Information and
  Modeling}}  \bibinfo{volume}{55} (\bibinfo{year}{2015}), \bibinfo{pages}{2324
  -- 2337}.
\newblock


\bibitem[Sun et~al\mbox{.}(2020a)]%
        {Sun2020InfoGraphUA}
\bibfield{author}{\bibinfo{person}{Fan-Yun Sun}, \bibinfo{person}{Jordan
  Hoffmann}, {and} \bibinfo{person}{Jian Tang}.}
  \bibinfo{year}{2020}\natexlab{a}.
\newblock \showarticletitle{InfoGraph: Unsupervised and Semi-supervised
  Graph-Level Representation Learning via Mutual Information Maximization}. In
  \bibinfo{booktitle}{\emph{ICLR}}.
\newblock


\bibitem[Sun et~al\mbox{.}(2020b)]%
        {Sun2020MultiStageSL}
\bibfield{author}{\bibinfo{person}{Ke Sun}, \bibinfo{person}{Zhanxing Zhu},
  {and} \bibinfo{person}{Zhouchen Lin}.} \bibinfo{year}{2020}\natexlab{b}.
\newblock \showarticletitle{Multi-Stage Self-Supervised Learning for Graph
  Convolutional Networks}. In \bibinfo{booktitle}{\emph{AAAI}}.
\newblock


\bibitem[Sun et~al\mbox{.}(2021)]%
        {Sun2021MoCLDM}
\bibfield{author}{\bibinfo{person}{Mengying Sun}, \bibinfo{person}{Jing Xing},
  \bibinfo{person}{Huijun Wang}, \bibinfo{person}{Bin Chen}, {and}
  \bibinfo{person}{Jiayu Zhou}.} \bibinfo{year}{2021}\natexlab{}.
\newblock \showarticletitle{MoCL: Data-driven Molecular Fingerprint via
  Knowledge-aware Contrastive Learning from Molecular Graph}. In
  \bibinfo{booktitle}{\emph{SIGKDD}}.
\newblock


\bibitem[Tang et~al\mbox{.}(2015)]%
        {Tang2015LINELI}
\bibfield{author}{\bibinfo{person}{Jian Tang}, \bibinfo{person}{Meng Qu},
  \bibinfo{person}{Mingzhe Wang}, \bibinfo{person}{Ming Zhang},
  \bibinfo{person}{Jun Yan}, {and} \bibinfo{person}{Qiaozhu Mei}.}
  \bibinfo{year}{2015}\natexlab{}.
\newblock \showarticletitle{LINE: Large-scale Information Network Embedding}.
  In \bibinfo{booktitle}{\emph{WWW}}.
\newblock


\bibitem[Veli{\v{c}}kovi{\'c} et~al\mbox{.}(2017)]%
        {velivckovic2017graph}
\bibfield{author}{\bibinfo{person}{Petar Veli{\v{c}}kovi{\'c}},
  \bibinfo{person}{Guillem Cucurull}, \bibinfo{person}{Arantxa Casanova},
  \bibinfo{person}{Adriana Romero}, \bibinfo{person}{Pietro Lio}, {and}
  \bibinfo{person}{Yoshua Bengio}.} \bibinfo{year}{2017}\natexlab{}.
\newblock \showarticletitle{Graph attention networks}.
\newblock \bibinfo{journal}{\emph{arXiv preprint arXiv:1710.10903}}
  (\bibinfo{year}{2017}).
\newblock


\bibitem[Wasserman and Faust(1994)]%
        {wasserman1994social}
\bibfield{author}{\bibinfo{person}{Stanley Wasserman} {and}
  \bibinfo{person}{Katherine Faust}.} \bibinfo{year}{1994}\natexlab{}.
\newblock \showarticletitle{Social network analysis: Methods and applications}.
\newblock  (\bibinfo{year}{1994}).
\newblock


\bibitem[Welling and Kipf(2016)]%
        {welling2016semi}
\bibfield{author}{\bibinfo{person}{Max Welling} {and} \bibinfo{person}{Thomas~N
  Kipf}.} \bibinfo{year}{2016}\natexlab{}.
\newblock \showarticletitle{Semi-supervised classification with graph
  convolutional networks}. In \bibinfo{booktitle}{\emph{ICLR}}.
\newblock


\bibitem[Wu et~al\mbox{.}(2018)]%
        {wu2018moleculenet}
\bibfield{author}{\bibinfo{person}{Zhenqin Wu}, \bibinfo{person}{Bharath
  Ramsundar}, \bibinfo{person}{Evan~N Feinberg}, \bibinfo{person}{Joseph
  Gomes}, \bibinfo{person}{Caleb Geniesse}, \bibinfo{person}{Aneesh~S Pappu},
  \bibinfo{person}{Karl Leswing}, {and} \bibinfo{person}{Vijay Pande}.}
  \bibinfo{year}{2018}\natexlab{}.
\newblock \showarticletitle{MoleculeNet: a benchmark for molecular machine
  learning}.
\newblock \bibinfo{journal}{\emph{Chemical science}} \bibinfo{volume}{9},
  \bibinfo{number}{2} (\bibinfo{year}{2018}), \bibinfo{pages}{513--530}.
\newblock


\bibitem[Xia et~al\mbox{.}(2019)]%
        {xia2019random}
\bibfield{author}{\bibinfo{person}{Feng Xia}, \bibinfo{person}{Jiaying Liu},
  \bibinfo{person}{Hansong Nie}, \bibinfo{person}{Yonghao Fu},
  \bibinfo{person}{Liangtian Wan}, {and} \bibinfo{person}{Xiangjie Kong}.}
  \bibinfo{year}{2019}\natexlab{}.
\newblock \showarticletitle{Random walks: A review of algorithms and
  applications}.
\newblock \bibinfo{journal}{\emph{IEEE Transactions on Emerging Topics in
  Computational Intelligence}} \bibinfo{volume}{4}, \bibinfo{number}{2}
  (\bibinfo{year}{2019}), \bibinfo{pages}{95--107}.
\newblock


\bibitem[Xia et~al\mbox{.}(2022)]%
        {Xia2022TowardsEA}
\bibfield{author}{\bibinfo{person}{Jun Xia}, \bibinfo{person}{Jiangbin Zheng},
  \bibinfo{person}{Cheng Tan}, \bibinfo{person}{Ge Wang}, {and}
  \bibinfo{person}{Stan~Z. Li}.} \bibinfo{year}{2022}\natexlab{}.
\newblock \showarticletitle{Towards Effective and Generalizable Fine-tuning for
  Pre-trained Molecular Graph Models}.
\newblock \bibinfo{journal}{\emph{bioRxiv}} (\bibinfo{year}{2022}).
\newblock


\bibitem[Xu et~al\mbox{.}(2021)]%
        {Xu2021LearningGA}
\bibfield{author}{\bibinfo{person}{Hongteng Xu}, \bibinfo{person}{Peilin Zhao},
  \bibinfo{person}{Junzhou Huang}, {and} \bibinfo{person}{Dixin Luo}.}
  \bibinfo{year}{2021}\natexlab{}.
\newblock \showarticletitle{Learning Graphon Autoencoders for Generative Graph
  Modeling}.
\newblock \bibinfo{journal}{\emph{ArXiv}}  \bibinfo{volume}{abs/2105.14244}
  (\bibinfo{year}{2021}).
\newblock


\bibitem[Xu et~al\mbox{.}(2019)]%
        {xu2018powerful}
\bibfield{author}{\bibinfo{person}{Keyulu Xu}, \bibinfo{person}{Weihua Hu},
  \bibinfo{person}{Jure Leskovec}, {and} \bibinfo{person}{Stefanie Jegelka}.}
  \bibinfo{year}{2019}\natexlab{}.
\newblock \showarticletitle{How Powerful are Graph Neural Networks?}. In
  \bibinfo{booktitle}{\emph{ICLR}}.
\newblock


\bibitem[Yang and Leskovec(2012)]%
        {Yang2012DefiningAE}
\bibfield{author}{\bibinfo{person}{Jaewon Yang} {and} \bibinfo{person}{Jure
  Leskovec}.} \bibinfo{year}{2012}\natexlab{}.
\newblock \showarticletitle{Defining and evaluating network communities based
  on ground-truth}.
\newblock \bibinfo{journal}{\emph{Knowledge and Information Systems}}
  \bibinfo{volume}{42} (\bibinfo{year}{2012}), \bibinfo{pages}{181--213}.
\newblock


\bibitem[You et~al\mbox{.}(2021)]%
        {you2021graph}
\bibfield{author}{\bibinfo{person}{Yuning You}, \bibinfo{person}{Tianlong
  Chen}, \bibinfo{person}{Yang Shen}, {and} \bibinfo{person}{Zhangyang Wang}.}
  \bibinfo{year}{2021}\natexlab{}.
\newblock \showarticletitle{Graph contrastive learning automated}. In
  \bibinfo{booktitle}{\emph{ICML}}. PMLR, \bibinfo{pages}{12121--12132}.
\newblock


\bibitem[You et~al\mbox{.}(2020a)]%
        {you2020graph}
\bibfield{author}{\bibinfo{person}{Yuning You}, \bibinfo{person}{Tianlong
  Chen}, \bibinfo{person}{Yongduo Sui}, \bibinfo{person}{Ting Chen},
  \bibinfo{person}{Zhangyang Wang}, {and} \bibinfo{person}{Yang Shen}.}
  \bibinfo{year}{2020}\natexlab{a}.
\newblock \showarticletitle{Graph contrastive learning with augmentations}. In
  \bibinfo{booktitle}{\emph{NeurIPS}}, Vol.~\bibinfo{volume}{33}.
  \bibinfo{pages}{5812--5823}.
\newblock


\bibitem[You et~al\mbox{.}(2020b)]%
        {You2020WhenDS}
\bibfield{author}{\bibinfo{person}{Yuning You}, \bibinfo{person}{Tianlong
  Chen}, \bibinfo{person}{Zhangyang Wang}, {and} \bibinfo{person}{Yang Shen}.}
  \bibinfo{year}{2020}\natexlab{b}.
\newblock \showarticletitle{When Does Self-Supervision Help Graph Convolutional
  Networks?}. In \bibinfo{booktitle}{\emph{PMLR}}, Vol.~\bibinfo{volume}{119}.
  \bibinfo{pages}{10871--10880}.
\newblock


\bibitem[Zhang et~al\mbox{.}(2019b)]%
        {Zhang2019OAGTL}
\bibfield{author}{\bibinfo{person}{Fanjin Zhang}, \bibinfo{person}{Xiao Liu},
  \bibinfo{person}{Jie Tang}, \bibinfo{person}{Yuxiao Dong},
  \bibinfo{person}{Peiran Yao}, \bibinfo{person}{Jie Zhang},
  \bibinfo{person}{Xiaotao Gu}, \bibinfo{person}{Yan Wang},
  \bibinfo{person}{Bin Shao}, \bibinfo{person}{Rui Li}, {and}
  \bibinfo{person}{Kuansan Wang}.} \bibinfo{year}{2019}\natexlab{b}.
\newblock \showarticletitle{OAG: Toward Linking Large-scale Heterogeneous
  Entity Graphs}. In \bibinfo{booktitle}{\emph{SIGKDD}}.
\newblock


\bibitem[Zhang et~al\mbox{.}(2019a)]%
        {Zhang2019ProNEFA}
\bibfield{author}{\bibinfo{person}{Jie Zhang}, \bibinfo{person}{Yuxiao Dong},
  \bibinfo{person}{Yan Wang}, \bibinfo{person}{Jie Tang}, {and}
  \bibinfo{person}{Ming Ding}.} \bibinfo{year}{2019}\natexlab{a}.
\newblock \showarticletitle{ProNE: Fast and Scalable Network Representation
  Learning}. In \bibinfo{booktitle}{\emph{IJCAI}}.
\newblock


\bibitem[Zhang et~al\mbox{.}(2022)]%
        {Zhang2022FineTuningGN}
\bibfield{author}{\bibinfo{person}{Jiying Zhang}, \bibinfo{person}{Xi Xiao},
  \bibinfo{person}{Long-Kai Huang}, \bibinfo{person}{Yu Rong}, {and}
  \bibinfo{person}{Yatao Bian}.} \bibinfo{year}{2022}\natexlab{}.
\newblock \showarticletitle{Fine-Tuning Graph Neural Networks via Graph
  Topology induced Optimal Transport}. In \bibinfo{booktitle}{\emph{IJCAI}}.
\newblock


\bibitem[Zhang et~al\mbox{.}(2021a)]%
        {Zhang2021MotifbasedGS}
\bibfield{author}{\bibinfo{person}{Zaixin Zhang}, \bibinfo{person}{Qi Liu},
  \bibinfo{person}{Hao Wang}, \bibinfo{person}{Chengqiang Lu}, {and}
  \bibinfo{person}{Chee-Kong Lee}.} \bibinfo{year}{2021}\natexlab{a}.
\newblock \showarticletitle{Motif-based Graph Self-Supervised Learning for
  Molecular Property Prediction}. In \bibinfo{booktitle}{\emph{NeurIPS}}.
\newblock


\bibitem[Zhang et~al\mbox{.}(2021b)]%
        {zhang2021motif}
\bibfield{author}{\bibinfo{person}{Zaixi Zhang}, \bibinfo{person}{Qi Liu},
  \bibinfo{person}{Hao Wang}, \bibinfo{person}{Chengqiang Lu}, {and}
  \bibinfo{person}{Chee-Kong Lee}.} \bibinfo{year}{2021}\natexlab{b}.
\newblock \showarticletitle{Motif-based graph self-supervised learning for
  molecular property prediction}. In \bibinfo{booktitle}{\emph{NeurIPS}},
  Vol.~\bibinfo{volume}{34}. \bibinfo{pages}{15870--15882}.
\newblock


\bibitem[Zhao et~al\mbox{.}(2021)]%
        {Zhao2021DataAF}
\bibfield{author}{\bibinfo{person}{Tong Zhao}, \bibinfo{person}{Yozen Liu},
  \bibinfo{person}{Leonardo Neves}, \bibinfo{person}{Oliver~J. Woodford},
  \bibinfo{person}{Meng Jiang}, {and} \bibinfo{person}{Neil Shah}.}
  \bibinfo{year}{2021}\natexlab{}.
\newblock \showarticletitle{Data Augmentation for Graph Neural Networks}. In
  \bibinfo{booktitle}{\emph{AAAI}}.
\newblock


\bibitem[Zhu et~al\mbox{.}(2021)]%
        {Zhu2021TransferLO}
\bibfield{author}{\bibinfo{person}{Qi Zhu}, \bibinfo{person}{Yidan Xu},
  \bibinfo{person}{Haonan Wang}, \bibinfo{person}{Chao Zhang},
  \bibinfo{person}{Jiawei Han}, {and} \bibinfo{person}{Carl Yang}.}
  \bibinfo{year}{2021}\natexlab{}.
\newblock \showarticletitle{Transfer Learning of Graph Neural Networks with
  Ego-graph Information Maximization}. In \bibinfo{booktitle}{\emph{NeurIPS}}.
\newblock


\bibitem[Zhu et~al\mbox{.}(2020)]%
        {Zhu2020DeepGC}
\bibfield{author}{\bibinfo{person}{Yanqiao Zhu}, \bibinfo{person}{Yichen Xu},
  \bibinfo{person}{Feng Yu}, \bibinfo{person}{Q. Liu}, \bibinfo{person}{Shu
  Wu}, {and} \bibinfo{person}{Liang Wang}.} \bibinfo{year}{2020}\natexlab{}.
\newblock \showarticletitle{Deep Graph Contrastive Representation Learning}.
\newblock \bibinfo{journal}{\emph{ArXiv}}  \bibinfo{volume}{abs/2006.04131}
  (\bibinfo{year}{2020}).
\newblock


\end{thebibliography}

\appendix
\newpage

\section{Proofs}\label{app:proof}
\subsection{Proof of Theorem~\ref{thero:emb}}\label{proof:emb}

\noindent \textit{\textbf{Proof:}}
We concentrate on the center node's embedding obtained from a $K$-layer GCN with 1-hop polynomial filter $\Phi (L)=I d-L$ whch is the common used GNN model. We denote the embedding of node $x_i \forall i=1, \cdots, n$ from a node-wise view in the final layer of the GCN as
$$
z_i^{(K)}=\sigma\left(\sum_{j \in \mathcal{N}\left(x_i\right)} a_{i j} z_j^{(K-1)^T} \theta^{(K)}\right) \in \mathbb{R}^d,
$$
where $a_{i j}=[\Phi (L)]_{i j} \in \mathbb{R}$ the weighted link between node $i$ and $j$; and $\theta^{(K)} \in \mathbb{R}^{d \times d}$ is the weight for the $K$ th layer sharing across nodes. Then $\theta=\left\{\theta^{(\ell)}\right\}_{\ell=1}^K$. We may denote $z_i^{(\ell)} \in \mathbb{R}^d$ similarly for $\ell=1, \cdots, K-1$, and $z_i^0=x_i \in \mathbb{R}^d$ as the node feature of center node $x_i$. With the assumption of GCN in the statement, we consider that only the k-hop ego-graph $g_i$ centered at $x_i$ is needed to compute $z_i^{(K)}$ for any $i=1, \cdots, n$ instead of the whole of $G$.

Denote $L_{g_i}$ as the out-degree normalised graph Laplacian of $g_i$, which is defined with respect to the direction from leaves  to centre node in $g_i$. We write the $\ell$ th layer embedding in following form
\begin{equation}
\left[z_i^{(\ell)}\right]_{K-\ell+1}=\sigma\left(\left[\Phi\left(L_{g_i}\right)\right]_{K-\ell+1}\left[z_i^{(\ell-1)}\right]_{K-\ell+1} \theta^{(\ell)}\right).
\end{equation}

Assume that $\forall i$, $\max _{\ell}\left\|z_i^{(\ell)}\right\|_2 \leq c_z $, and $\max _{\ell}\left\|\theta^{(\ell)}\right\|_2 \leq c_\theta$. Suppose that the activation function $\sigma$ is $\kappa_\sigma$-Lipschitz function Then, for $\ell=1, \cdots, K-1$, we have
\begin{equation}
\begin{aligned}
&\left\|\left[z_{i}^{(\ell)}\right]_{K-\ell}-\left[z_{i^{\prime}}^{(\ell)}\right]_{K-\ell}\right\|_2 \\
&\leq\|[\sigma\left(\left[\Phi\left(L_{g_i}\right)\right]_{K-\ell+1}\left[z_i^{(\ell-1)}\right]_{K-\ell+1} \theta^{(\ell)}\right) \\
&-\sigma\left(\left[\Phi\left(L_{g_{i^{\prime}}}\right)\right]_{K-\ell+1}\left[z_{i^{\prime}}^{(\ell-1)}\right]_{K-\ell+1} \theta^{(\ell)}\right)]_{K-\ell}) \|_2 \\
& \leq \kappa_\sigma c_\theta\left\|\Phi\left(L_{g_i}\right)\right\|_2\left\|\left[z_i^{(\ell-1)}\right]_{K-\ell+1}-\left[z_{i^{\prime}}^{(\ell-1)}\right]_{K-\ell+1}\right\|_2 \\
&+\kappa_\sigma c_\theta c_z\left\|\Phi\left(L_{g_i}\right)-\Phi\left(L_{g_{i^{\prime}}}\right)\right\|_2 .
\end{aligned}
\end{equation}
Since $\left[\Phi\left(L_{g_i}\right)\right]_{K-\ell+1}$ is the principle submatrix of $\Phi\left(L_{g_i}\right)$. We equivalently write the above equation as $A_{\ell} \leq a A_{\ell-1}+b$, where $a$ and $b$ is the coefficient. And we have
\begin{equation}
\begin{aligned}
A_{\ell} &\leq  a A_{\ell-1}+b \\
&\leq  a^{2} A_{\ell-2}+b(1+a) \\
& \cdots\\
&\leq a^{\ell} A_1+\frac{a^{\ell}+1}{a-1} b.\\
\end{aligned}
\end{equation}

Therefore, for any $\ell=1, \cdots, K$, we have an upper bound for the hidden representation difference between $g_i$ and $g_i^{\prime}$ by substitute coefficient $a$ and $b$,
\begin{equation}
\begin{aligned}
\left\|\left[z_i^{(\ell)}\right]_{K-\ell}-\left[z_{i^{\prime}}^{(\ell)}\right]_{K-\ell}\right\|_2 & \leq\left(\kappa_\sigma c_\theta\right)^{\ell}\left\|\Phi\left(L_{g_i}\right)\right\|_2^{\ell}\left\|\left[x_i\right]-\left[x_{i^{\prime}}\right]\right\|_2 \\
& +\frac{\left(\kappa_\sigma c_\theta\right)^{\ell}\left\|\Phi\left(L_{g_i}\right)\right\|_2^{\ell}+1}{\kappa_\sigma c_\theta\left\|\Phi\left(L_{g_i}\right)\right\|_2-1} \kappa_\sigma c_\theta c_z\left\|\Phi\left(L_{g_i}\right)-\Phi\left(L_{g_{i^{\prime}}}\right)\right\|_2 .
\end{aligned}
\end{equation}
Specifically, for $\ell=K$, we obtain the upper bound for center node embedding $\left\|\left[z_i^{(K)}\right]_0-\left[z_{i^{\prime}}^{(K)}\right]_0\right\| \equiv$ $\left\|z_i-z_{i^{\prime}}\right\|$. Since the attribute of each node as a scalar 1, we therefore have $\left\|\left[x_i\right]-\left[x_{i^{\prime}}\right]\right\|_2 =0$. Suppose that $\forall i$, $\left\|\Phi\left(L_{g_i}\right)\right\|_2 \leq c_L$ 
because that the the graph Laplacians are normalised. Since $\Phi$ is a linear function for $L$, We have 
\begin{equation}\label{eq:divergence}
\begin{aligned}
\left\|z_i-z_{i^{\prime}}\right\|_2 & \leq\frac{\left(\kappa_\sigma c_\theta c_L\right)^K+1}{\kappa_\sigma c_\theta c_L-1} 
c_\theta c_z\left\|\Phi\left(L_{g_i}\right)-\Phi\left(L_{g_{i^{\prime}}}\right)\right\|_2 \\
& \leq \kappa\left\|L_{g_i}-L_{g_{i^{\prime}}}\right\|_2, 
\end{aligned}
\end{equation}
where $\kappa=\frac{\left(\kappa_\sigma c_\theta c_L\right)^K+1}{\kappa_\sigma c_\theta c_L-1} 
c_\theta c_z$.

Therefore, we sequentially compute the representation divergence between node $i$ and node $j^{\prime}$ in $G_\text{train}$ and $G_\text{down}$ respectively. By means of Eq.(\ref{eq:divergence}), we have
\begin{equation}
\begin{aligned}
\left\|e(G_\text{train})-e(G_\text{down})\right\|_2
& = \frac{1}{mn}  \sum_{i=1}^m \sum_{j=1}^n \left\| z_i-z_{j^{\prime}}\right\|_2\\
&\leq \frac{\kappa}{mn}  \sum_{i=1}^m \sum_{j=1}^n \left\|L_{g_i}-L_{g_{j^{\prime}}}\right\|_2. \\
\end{aligned}
\end{equation}

 \subsection{Proof of Theorem \ref{theor:dis-prob}}\label{proof:dis-pro}
We first show the following lemma, which would be used in the proof of Theorem \ref{theor:dis-prob}, 
\begin{lemma}\label{lemma:convergent}
(Proposition 11.32 in~\cite{lovasz2012large}) For every graphon $W$, generating a $W$-random graph $\mathbb{G}(n, W)$ for $n=1,2, \ldots$ we get a graph sequence such that $\mathbb{G}(n, W) \rightarrow W$ with probability 1.
\end{lemma} 
\noindent \textit{\textbf{Proof of Theorem \ref{theor:dis-prob} :}} Since we assume $\mathcal{G}_\text{down}$ can be generated from $f(\{\alpha_i\},\{B_i\})$ with  probability 1, according to Lemma~\ref{lemma:convergent}, we can have that $\mathcal{G}_\text{down} \rightarrow f(\{\alpha_i\},\{B_i\} )$ with  probability 1.
On the other hand, since $B_\text{down}$ is the  graphon fitted by  $\mathcal{G}_\text{down}$, which means $\mathcal{G}_\text{down}$ is convergent to graphon $B_\text{down}$ and we have $\mathcal{G}_\text{down} \rightarrow B_\text{down} $. Hence $B_\text{down}$ is equivalent to $f$, then dist($f, B_\text{down}$) =0.

\subsection{Proof of Theorem~\ref{theor-preserve}}
\noindent \textit{\textbf{Proof:}}
Given an arbitrary set of graphons $X=\{B_i\}$ and $i=1...k$ the 
 general form of the convex combination of  graphons in $X$ can be represented as:
\begin{equation}
f(\{\alpha_i\},\{B_i\}) = \sum _{i=1}^k \alpha_i B_i,\quad \quad
(\sum _{i=1}^k \alpha_i = 1,\alpha_i \geq 0).
\end{equation}
Next we prove that our generator preserves the properties of graphons.
the  convex combination of  graphons is still a bounded symmetric function $[0,1]^2 \to [0,1]$.
First, we prove  that the convex combination $f(\{\alpha_i\},\{B_i\})$ is still symmetric. Since $f(\{\alpha_i\},\{B_i\})$ is symmetric if and only if it satisfies that $f(\{\alpha_i\},\{B_i\}) = f(\{\alpha_i\},\{B_i\})^T$, we have the following derivations:
\begin{equation}
\begin{aligned}
    {f(\{\alpha_i\},\{B_i\})}^T&=
(\alpha_1B_1+...+ \alpha_kB_k
)^T\\
&= \alpha_1B_1^T+...+ \alpha_kB_k^T
\\&=\alpha_1B_1+...+ \alpha_kB_k = f(\{\alpha_i\},\{B_i\}).
\end{aligned}
\end{equation}
Then we prove that the convex combination $f(\{\alpha_i\},\{B_i\})$ is still in $[0,1]$
Let $B_{max}$ indicates the maximum $B_i$, meanwhile $B_{min}$ indicates the minimum $B_i$
\begin{equation}
\begin{aligned}
 f(\{\alpha_i\},\{B_i\}) = (\alpha_1B_1+...+ \alpha_kB_k
) \leq \sum_{i=1}^k{\alpha_i}B_{max} = B_{max} \leq 1.
\end{aligned}
\end{equation}
\begin{equation}
\begin{aligned}
f(\{\alpha_i\},\{B_i\}) = (\alpha_1B_1+...+ \alpha_kB_k
) \geq  \sum_{i=1}^k{\alpha_i}B_{min} \geq B_{min} \ge 0.
\end{aligned}
\end{equation}
% Hence we have proved that the convex combination of graphons preserve the properties of graphon and is still a bounded symmetric function $[0,1]^2 \to [0,1]$.
Thus, for a set of graphon basis $X$, the generator space $\Omega$ is the set of all convex combinations of basis elements in $X$, $\Omega$ is the convex hull of $X$.

\subsection{Proof of Theorem~\ref{theor:diff}}

We first show the following two lemmas, which would be used in the proof of  Theorem~\ref{theor:diff}.

The first lemma is known as counting lemma for graphons, provided in Lemma 10.23 in~\cite{lovasz2012large}.
\begin{lemma}\label{lemma:counting}
Let $F$ be a graph motif and let $B$, $B'$ be two graphon . Then
we have
\begin{equation}
\left|t(F, B)-t\left(F, B^{\prime}\right)\right| \leq |F|\left\|B-B^{\prime}\right\|_{\square}.
\end{equation}
\end{lemma}
\noindent \textit{\textbf{Proof of Lemma~\ref{lemma:counting}:}}
\begin{equation}
\begin{aligned}
& \left|t(F, B)-t\left(F, B^{\prime}\right)\right| \\
& =\left|\int\left(\prod_{u_{i} v_{i} \in E} B\left(u_{i}, v_{i}\right)-\prod_{u_{i} v_{i} \in E} B^{\prime}\left(u_{i}, v_{i}\right)\right) \prod_{v \in V} d v\right| \\
& \leq \sum_{i=1}^{|E|}\left|\int\left(\prod_{j=1}^{i-1} B^{\prime}\left(u_{j}, v_{j}\right)\left(B\left(u_{i}, v_{i}\right)-B^{\prime}\left(u_{i}, v_{i}\right)\right) \prod_{k=i+1}^{|E|} B\left(u_{k}, v_{k}\right)\right) \prod_{v \in V} d v\right|.
\end{aligned}
\end{equation}

Each absolute value term in the sum is bounded by the cut norm $\left\|B-B^{\prime}\right\|_{\square}$. When we fix all other irrelavant variables (everything except $u_{i}$ and $v_{i}$ for the $i$-th term), altogether implying that

\begin{equation}
\left|t(F, B)-t\left(F, B^{\prime}\right)\right| \leq |F|\left\|B-B^{\prime}\right\|_{\square}.
\end{equation}
\begin{lemma}\label{lemma:cut norm}
The cut norm  of a graphon $\|B\|_{\square}$ is defined as 
\begin{equation}
\|B\|_{\square}=\sup _{S, T \subseteq[0,1]}\left|\int_{S \times T} B\right|,
\end{equation}
where the supremum is taken over all measurable subsets $S$ and $T$.
Obviously, suppose $\alpha \in \mathbb{R}$, we have
\begin{equation}
\|\alpha B\|_{\square}=\sup _{S, T \subseteq[0,1]}\left|\int_{S \times T} \alpha B\right|=\sup _{S, T \subseteq[0,1]}\left|\alpha \int_{S \times T} B\right|=\alpha\|B\|_{\square}.
\end{equation}

\end{lemma}
\noindent \textit{\textbf{Proof of Theorem~\ref{theor:diff}:}} Based on the Lemma~\ref{lemma:counting} and Lemma~\ref{lemma:cut norm}, we have the following derivations.
The $a-$th element $B_a$  in graphon basis has its corresponding motif set. Each  motif $F_a$ is expected to be preserved and to exhibit similar frequency (\emph{i.e., homomorphism density}) in  $f(\{\alpha_i\},\{B_i\})$. 

Applying Lemma~\ref{lemma:counting}, we have the following derivations:
\begin{equation}
\begin{aligned}
&|t(F_a,f(\{\alpha_i\},\{B_i\}))-t(F_a,B_a)| \leq |F_a|||f(\{\alpha_i\},\{B_i\}) -B_a|| _\square\\
&|t(F_a,f(\{\alpha_i\},\{B_i\}))-t(F_a,B_a)| \leq |F_a| ||\sum _{b=1}^k\alpha_b B_b - \sum _{b=1}^k \alpha_b B_a||_\square\\
&|t(F_a,f(\{\alpha_i\},\{B_i\}))-t(F_a,B_a)| \leq |F_a| ||\sum _{b=1}^k\alpha_b (B_b -B_a)||_\square.
\end{aligned}
\end{equation}
Combining with the triangle inequality, we have:
\begin{equation}
|t(F_a,f(\{\alpha_i\},\{B_i\}))-t(F_a,B_a)|\leq \sum _{b=1,b\neq a}^k  |F_a| \alpha_b || B_b -B_a||_\square.
\end{equation}

where a=1...k, $|F|$ represents the number of nodes in motif $F$, and $||
\cdot||_\square$ is the cut norm.

\subsection{Proof of Theorem~\ref{thm:down}}

\begin{lemma}\label{lemma:random}
    (Corollary $10.4$ in~\cite{lovasz2012large}). Let $W$ be a graphon, $n \geq 1,0<\varepsilon<1$, and let $F$ be a simple graph, then the $W$-random graph $\mathbb{G}=\mathbb{G}(n, W)$ satisfies
\begin{equation}
    \mathrm{P}(|t(F, \mathbb{G})-t(F, W)|>\varepsilon) \leq 2 \exp \left(-\frac{\varepsilon^2 n}{8 \mathrm{v}(F)^2}\right).
\end{equation}

\end{lemma}

\noindent \textit{\textbf{Proof of Theorem~\ref{thm:down}:}}
Apply Lemma~\ref{lemma:random} in our setting, let graphon $B$ as $W$ Lemma~\ref{lemma:random}, we have 
\begin{equation}
\mathrm{P}(|t(F, \mathbb{G})-t(F, B)|>\varepsilon) \leq 2 \exp \left(-\frac{\varepsilon^2 n}{8 \mathrm{v}(F)^2}\right).
\end{equation}

\section{Datasets}\label{app:data}
\vpara{Datasets for node classification.}
For pre-training datasets, following~\cite{Qiu2020GCCGC},  we collect the graph data from Academia (NetRep)~\cite{Ritchie2016ASP}, DBLP (SNAP)~\cite{Yang2012DefiningAE} and DBLP (NetRep)~\cite{Ritchie2016ASP} as the academic domain data. For movie domain, we utilize the graph from IMDB~\cite{Ritchie2016ASP}. As for social domain, the graph data from Facebook~\cite{Ritchie2016ASP} and LiveJournal~\cite{Backstrom2006GroupFI} is leveraged.
For the downstream dataset, we use data from transportation, academic and web domains.
Specifically, we collect the datasets from H-Index~\cite{Zhang2019OAGTL} and Chamelon~\cite{Rozemberczki2021MultiscaleAN} for academic and web domains, respectively.
The datasets in transportation domain are collected from US-Airport and Europe-Airport~\cite{Ribeiro2017struc2vecLN}.
Detailed statistics of pre-training and downstream datasets for node classification is summarized in Table~\ref{tab:dataset}.

\begin{table*}[!ht]
\renewcommand\arraystretch{1.2}
\centering
 \resizebox{0.95\linewidth}{!}
 {
    \setlength\tabcolsep{2pt} 
    \begin{threeparttable}
    \begin{tabular}{ccccccccccc}
     \toprule
        & \multicolumn{1}{c}{Type}  &  \multicolumn{1}{c}{Name}  &  \multicolumn{1}{c}{$|{V}|$}  &  \multicolumn{1}{c}{$|{E}|$}  &  avg deg & deg var& avg clustering coef & density & assortativity coef & transitivity \\
     \midrule
     \multirow{6}*{\setlength\tabcolsep{1pt}\rotatebox{90}{\textbf{pre-training}}} & {\textcolor[rgb]{0.404,0.173,0.075}{\textbf{academic}}}  &  Academia &  137,969  &  739,384  &  5.36 & 10.11 & 1.42e-01 & 3.88e-05 & 8.39e-03 & 7.65e-02 \\
     & ~   &  DBLP(SNAP) & 317,080 & 2,099,732 & 6.62 & 10.01& 6.32e-01 & 2.09e-05 & 2.67e-01 & 3.06e-01 \\
     & ~   &  DBLP(NetRep) & 540,686  & 30,491,458 & 56.41 & 66.24& 8.02e-01 & 1.04e-04 & 5.10e-01 & 6.56e-01  \\
     & {\textcolor[rgb]{0.765,0.4078,0.145}{\textbf{social}}}  &  Facebook  & 3,097,165 & 47,334,788 & 15.28 & 45.17 & 9.70e-02& 4.93e-06 & -5.57e-02 & 4.77e-02  \\
     & ~   &  LiveJournal & 4,843,953 & 85,691,368 & 17.69 & 52.02 & 2.74e-01& 3.65e-06 & 2.10e-02 & 1.18e-01 \\
     & {\textcolor[rgb]{0.878,0.690,0.169}{\textbf{movie}}}  & IMDB &  896,305  &  3,782,447 &  8.44 & 17.27& 5.79e-05 & 9.42e-06 & -5.30e-02 & 8.08e-05 \\
     \midrule
     \multirow{2}*{\rotatebox{90}{\textbf{downstream}}} &
    {\textcolor[rgb]{0.404,0.173,0.075}{\textbf{academic}}}  &  H-index & 5000 & 44,020 & 17.61& 31.91& 7.10e-01 & 3.52e-03 & 1.18e-01 & 5.66-01 \\
     & {\textcolor[rgb]{0.9,0.2,0.3294}{\textbf{web}}} &  Chameleon  & 2277  & 36,101 & 27.58 & 46.40 & 4.81e-01& 1.21e-02 & -1.99e-01 & 3.14e-01 \\
     & {\textcolor[rgb]{0.075,0.275,0.514}{\textbf{transportation}}}   & US-Airport & 1,190  & 13,599 & 22.86 & 40.45& 5.01e-01 & 1.92e-02 & 3.13e-02 & 4.26e-01 \\ 
     & ~ & Europe-Airport & 399 & 5,995 & 30.05 & 34.65& 5.39e-01 & 7.55e-02 & -2.25e-01 & 3.34e-01 \\
    \bottomrule
    \end{tabular}
    \end{threeparttable}
 }
\caption{The statistics of pre-training and downstream datasets (avg deg: average degree; deg var: degree variance; coef: coefficient) for node classification.}
\label{tab:dataset}
\vspace{-0.15in}
\end{table*}
\vpara{Datasets for graph classification.}
For pre-training datasets, to enrich the follow-up experimental analysis, we use scaffold split to partition the ZINC15 into five datasets (ZINC15-0,ZINC15-1,ZINC15-2,ZINC15-3 and ZINC15-4) according to their scaffolds~\cite{hu2019strategies}, such that the scaffolds are different in each dataset.
{For downstream datasets, we use BACE, BBBP, MUV, HIV, and ClinTox provided in~\cite{wu2018moleculenet}. }
BACE provides quantitative ($IC_{50}$) and qualitative (binary label) binding results for a set of inhibitors of human $\beta$-secretase 1, which merged a collection of 1522 compounds with their 2D structures and binary labels in MoleculeNet. BBBP includes binary labels for over 2000 compounds on their permeability properties. MUV contains 17 tasks for around 90 thousand compounds, designed for validation of virtual screening techniques. HIV was introduced to test the ability to inhibit HIV replication for over 40,000 compounds. ClinTox includes two classification tasks for 1491 drug compounds with known chemical structures.
Detailed statistics of pre-training and downstream datasets for graph classification is summarized in Table~\ref{tab:dataset_graph}.

\begin{table*}[!ht]
\renewcommand\arraystretch{1.2}
% \vspace{-0.1in}
\centering
 \resizebox{0.95\linewidth}{!}
 {
    \setlength\tabcolsep{2pt} 
    \begin{threeparttable}
    \begin{tabular}{ccccccccccc}
     \toprule
        &  \multicolumn{1}{c}{Name}  & avg node & avg edge & avg deg & deg var & density & closeness & assortativity coef & transitivity & avg clustering coef \\
     \midrule
     \multirow{6}*{\setlength\tabcolsep{1pt}\rotatebox{90}{\textbf{pre-training}}} &  ZINC15-0 &  26.95 & 29.24 & 2.16 & 0.70 & 0.08 & 0.19 & -0.31 & 4.22e-03 & 4.43e-03\\
     & ZINC15-1 & 25.69 & 27.67 & 2.14 & 0.71 & 0.09 & 0.20 & -0.33 & 3.08e-03 & 3.38e-03\\
     & ZINC15-2 & 26.86 & 29.21 & 2.17 & 0.69 & 0.08 & 0.19 & -0.31 & 3.66e-03 & 4.07e-03 \\
     & ZINC15-3 & 26.91 & 29.19 & 2.16 & 0.70 & 0.08 & 0.19 & -0.31 & 3.30e-03 & 3.66e-03\\
     & ZINC15-4 & 26.83 & 29.18 & 2.17 & 0.69 & 0.08 & 0.19 & -0.31 & 3.71e-02 & 4.09e-03\\
     \midrule
     \multirow{2}*{\rotatebox{90}{\textbf{downstream}}} &
        BACE & 34.08 & 36.85 & 2.16 & 0.76 & 0.07 & 0.17 & -0.36 & 6.10e-03 & 6.54e-03\\
     & BBBP & 24.06 & 25.95 & 2.13 & 0.75 & 0.11 & 0.24 & -0.27 & 2.21e-03 & 2.56e-03 \\
     & MUV & 24.23 & 26.27 & 2.16 & 0.68 & 0.09 & 0.20 & -0.27 & 8.75e-04 & 1.01e-03 \\ 
     & HIV & 25.51 & 27.46 & 2.14 & 0.74 & 0.10 & 0.22 & -0.26 & 2.11e-03 & 1.96e-03 \\
     & ClinTox & 26.15 & 27.88 & 2.10 & 0.77 & 0.11 & 0.23 & -0.34 & 2.98e-03 & 3.21e-03 \\
    \bottomrule
    \end{tabular}
    \end{threeparttable}
 }
\caption{The statistics of pre-training and downstream datasets (avg deg: average degree; deg var: degree variance; coef: coefficient) for graph classification. All these datasets are from molecular domain.}
% \JR{avg closeness?}}
\label{tab:dataset_graph}
\vspace{-0.15in}
\end{table*}

\section{Additional Results}

\subsection{Results of Pre-training Feasibility }\label{app:graph_class_res}
\begin{figure*}[!h]
 \centering\setlength{\abovecaptionskip}{0.15cm}{
 \includegraphics[width=\linewidth]
 {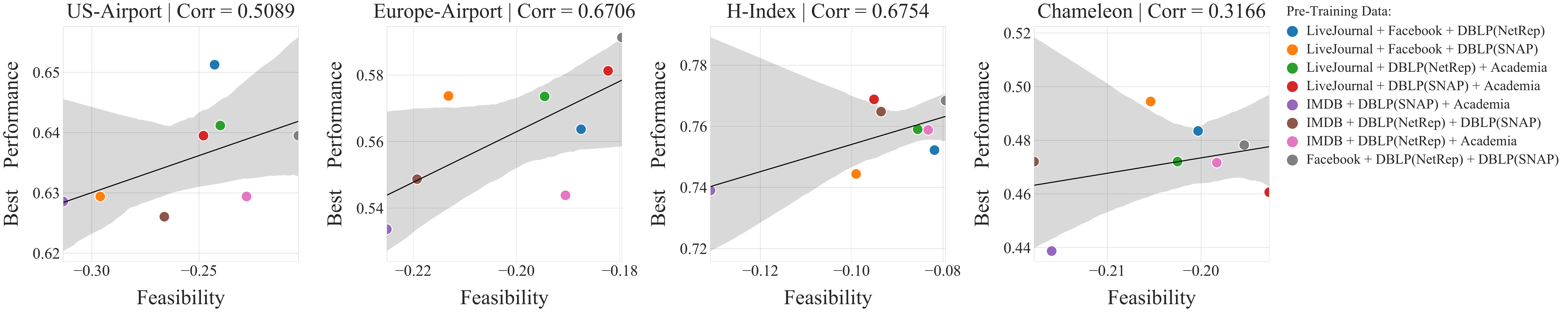}}
\caption{Pre-training feasibility vs. the best downstream performance on node classification when the selection buget is 3.}
\label{fig:corre_node_3}
\vspace{-0.1in}
\end{figure*}

\begin{figure*}[!h]
 \centering\setlength{\abovecaptionskip}{0.15cm}{\includegraphics[width=\linewidth]{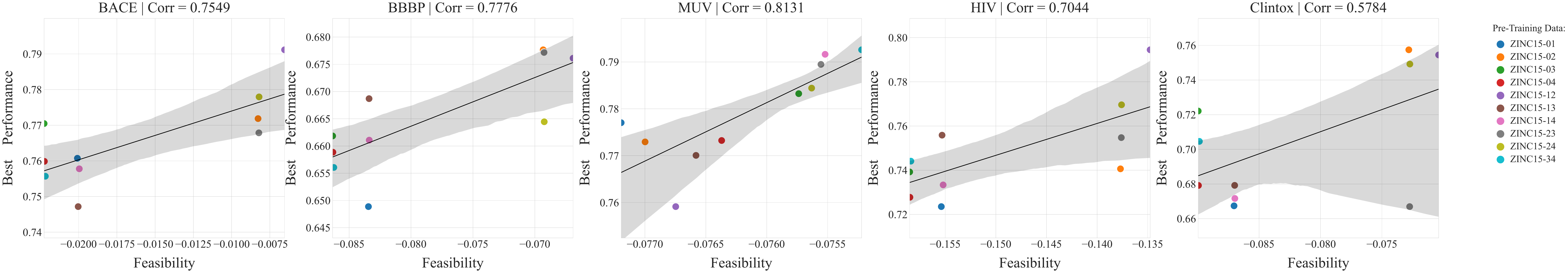}}
 \centering{\includegraphics[width=\linewidth]{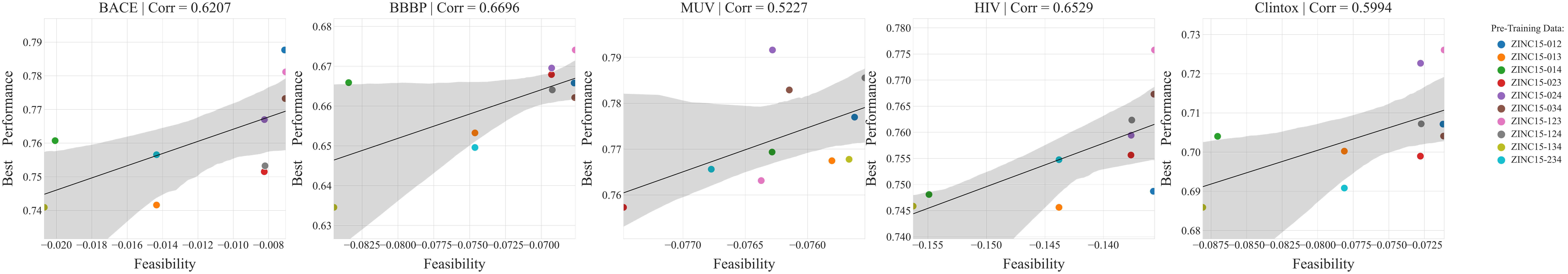}}
\caption{Pre-training feasibility vs. best downstream performance on graph classification when the selection buget is 2 and 3. }
\vspace{-0.1in}
\label{fig:corre_graph}
\end{figure*}

We here give additional plots to show the detailed correlation between estimated pre-training feasibility and the best downstream performance.
Figure~\ref{fig:corre_node_3} shows the results on node classification  when the selection budget is  3.
Figure~\ref{fig:corre_graph} shows the results on graph classification when the selection budget is 2 and 3 respectively.
We find that in all cases, there is a strong positive correlation between
pre-training feasibility and the best downstream performance.

\begin{table*}[h]
\centering
\resizebox{\linewidth}{!}{
\begin{tabular}{lcccccccccccc}
\toprule[1pt]
\multicolumn{1}{c}{\multirow{2}{*}{}} & \multicolumn{6}{c}{$N=2$}      & \multicolumn{6}{c}{$N=3$}    \\
\cmidrule[1pt](r){2-7}\cmidrule[1pt](r){8-13} \multicolumn{1}{c}{}                         & BACE  & BBBP  & MUV   & HIV   & ClinTox & Rank & BACE  & BBBP  & MUV   & HIV   & ClinTox & Rank \\ \bottomrule[1pt]
\cellcolor{ggray!90}All Datasets                                      & \cellcolor{ggray!90}74.86 & \cellcolor{ggray!90}62.67 & \cellcolor{ggray!90}73.80 & \cellcolor{ggray!90}68.31 & \cellcolor{ggray!90}60.58   & \cellcolor{ggray!90}-    & \cellcolor{ggray!90}74.86 & \cellcolor{ggray!90}62.67 & \cellcolor{ggray!90}73.80 & \cellcolor{ggray!90}68.31 & \cellcolor{ggray!90}60.58  & \cellcolor{ggray!90}-    \\
\toprule
Graph Statistics                             &  71.52&  58.47 & 69.42 & 70.31    &62.99 & 2    & 65.95&  62.36 & 68.98 & 68.83    &59.86 & 5    \\
EGI                                          & 68.46 & \textbf{65.65}&  69.42 & 68.29  &  \textbf{65.21}  & 3    & 68.48 &  61.00 & 68.98  &68.83    &60.35 & 3    \\
Clustering Coefficient                       &  71.52 & 60.06 & 69.42 & 70.31  &  58.91  & 5    &65.95 & 61.00 & 68.98&  68.83 &   60.35   & 6    \\
Spectrum of Graph Laplacian                  & 69.43&  \textbf{65.65}  &68.57 & 68.88  &  59.82  & 6    & 70.21  &61.00 & 69.14 & 69.18    &\textbf{63.02}& 2    \\
Betweenness Centrality           &  71.52 & 63.20 & 69.42 & \textbf{71.72} &   58.91 & 4    & 67.08  &\textbf{67.41} & 68.98  &68.83  &  60.35  & 4    \\ \midrule
W2PGNN                                       & \textbf{73.33} & 65.46  &\textbf{74.17} & 70.69  &  64.21   & 1    & \textbf{73.54} & 65.02  &\textbf{72.49}  &\textbf{71.18}&    62.26 & 1 \\ \bottomrule[1pt]   
\end{tabular}
}
\caption{Graph classification results when performing pre-training on different selected pre-training data. We also provide the results of using all pre-training data without selection for your reference (see ``All Datasets'' in the table).}
\label{tab:graph_classification_data_selection_results}
% \vspace{-0.2in}
\end{table*}
% \vspace{-0.25in}
\subsection{Results of Pre-Training Data Selection}\label{app:graph_class_res2}
Table~\ref{tab:graph_classification_data_selection_results} shows the results of pre-training data selection on graph classification tasks. 
The backbone pre-training model used here is GraphCL~\cite{you2020graph}.
We can see that the pre-training data selected by W2PGNN
ranks the first, which suggests that the effectiveness of our strategy on the graph classification task is still significant.

\vspace{-0.05in}
\section{Computation Complexity Analysis}\label{app:complexity}
We show the time complexity of W2PNN and the traditional  solution.
Suppose that we have $n_1$  and $n_2$ (sub)graphs sampled from pre-training data and  downstream data respectively. Denote $|V|$ and $|E|$ as the average number of nodes and edges per (sub)graph.

The time complexity of W2PGNN consists of three components: 
computation of three graphon base ($\{B_i\}_\text{topo}$, $\{B_i\}_\text{domain}$, $\{B_i\}_\text{integr}$), graphon estimation of downstream data (\emph{i.e.}, $B_\text{down}$), computation of GW distance (\emph{i.e.}, $\operatorname{dist}(\cdot, \cdot)$):
(1) Among the estimation of three graphon base, the topological feature extraction is the most time-consuming one. It mainly involves the topological feature extractor, K-means clustering and graphon estimation of pre-training data, which costs {$O(n_1|E|)$} (which is taken as the complexity of the most costly property closeness)~\cite{freeman2002centrality}, $O(n_1)$~\cite{Kaiser2008MeanCC} and $O(n_1|V|^2)$~\cite{channarond2012classification}, respectively. The domain and integrated graphon basis only include the graphon estimation of pre-training data and cost $O(n_1|V|^2)$~\cite{channarond2012classification}.
(2) The graphon estimation of downstream data costs $O(n_2|V|^2)$~\cite{channarond2012classification}.
(3) The computation of GW distance costs $O(|V|^3)$ ~\cite{peyre2016gromov}, which can be ignored because we have $|V| \ll n_1+n_2$
(For node-level transferable patterns, extracting the ego-network of sampled nodes via breadth first search costs $O((n_1+n_2)(|V|+|E|))$, which can be ignored).
So the overall time complexity of W2PGNN is $O((n_1+n_2)|V|^2)$.

For comparison, traditional solution make $l_1 \times l_2$ ``pre-train and fine-tune'' attempts, if there are $l_1$ pre-training models and $l_2$ fine-tuning strategies. Suppose the batch size of pre-training as $b$ and the representation dimension as $d$. The time complexity of each pre-training model (taking the most general graph pre-training model GCC~\cite{Qiu2020GCCGC} as example) is typically from data augmentation, GNN encoder and  contrastive loss, which costs $O\left(n_1|V|^3\right)$ (subgraphs sampled from random walk with restarts as augmentation)~\cite{xia2019random}, {$O\left(n_1|E|d\right)$} (GIN as graph encoder) and $O\left(n_1bd\right)$~\cite{li2022let}, respectively. 
(Note that other data augmentations like dropping nodes cost $O(|V|^2)$, but they cannot achieve good performance on node classification in our pre-training experiments.)
The time complexity of each fine-tuning strategy involves the inference of pre-trained model  $O\left(n_2 (|V|^3 + |E|d)\right)$ and downstream predictor  $O\left(n_2d\right)$ (which can be ignored), under the simple freezing mode.
% (\emph{i.e}, the pre-trained model without any changes in parameters is directly applied to the downstream tasks via learnable downstream predictor).
Thus the overall time complexity of traditional solution is 
 $O\left(l_1l_2((n_1+n_2)(|V|^3 + |E|d)+n_1bd)\right)$.

\end{document}